\documentclass[10pt,twocolumn,letterpaper]{article}

\usepackage[pagenumbers]{cvpr} 

\usepackage{colortbl}
\usepackage{booktabs}

\usepackage{multirow}
\usepackage{listings}
\usepackage{algorithm}
\usepackage{algpseudocode}
\usepackage{tikz}
\usepackage[most]{tcolorbox}

\definecolor{cvprblue}{rgb}{0.21,0.49,0.74}
\usepackage[pagebackref,breaklinks,colorlinks,allcolors=cvprblue]{hyperref}

\def\confName{CVPR}
\def\confYear{2027}

\title{EnRealWorld: 4D World Creation as Agentic Engineering}

\author{%
  Zihan Guo\textsuperscript{1,*,\textdagger},
  Yuanyi Song\textsuperscript{2,*},
  Yuan Yuan\textsuperscript{2},
  Zicai Cui\textsuperscript{2},
  Chenxi Zeng\textsuperscript{3},
  Shuai Shao\textsuperscript{2},
  Haoyuan Li\textsuperscript{1},\\[4pt]
  Qihang Cao\textsuperscript{5},
  Zunnan Xu\textsuperscript{5},
  Dongzhan Zhou\textsuperscript{4},
  Ming Zhou\textsuperscript{4},
  Jianghao Lin\textsuperscript{2,\textdaggerdbl},
  Yang Li\textsuperscript{2,\textdaggerdbl},
  Weinan Zhang\textsuperscript{2,\textdaggerdbl}\\[8pt]
  \textsuperscript{1}\,Sun Yat-sen University\quad
  \textsuperscript{2}\,Shanghai Jiao Tong University\quad
  \textsuperscript{3}\,Jilin University\\[3pt]
  \textsuperscript{4}\,Shanghai Artificial Intelligence Laboratory\quad
  \textsuperscript{5}\,Nanyang Technological University
}

\begin{document}
\maketitle
{%
  \renewcommand{\thefootnote}{\fnsymbol{footnote}}%
  \footnotetext[1]{Equal contribution. The co-first authors agree that each
    may list their own name first when presenting this work.}%
  \footnotetext[2]{Project lead. \nolinkurl{guozh29@mail2.sysu.edu.cn}.}%
  \footnotetext[3]{Co-corresponding authors.}%
}

\begin{abstract}
Interactive 4D worlds provide controllable environments for Physical AI, but visual realism and successful demonstration replay do not ensure reliable behavior under sustained interaction or external intervention. Creating executable worlds from heterogeneous inputs requires clarifying requirements, coordinating geometry and dynamics, and obtaining diagnostic feedback. We introduce EnRealWorld, which formalizes 4D world creation as agentic engineering. It is a sequential decision process that couples specification refinement, world construction, proactive investigation, and cross-task experience accumulation. The Manager maintains a testable, implementation-independent specification, and the Engineer builds and revises candidate worlds. The Researcher independently selects observations and interventions to uncover hidden failures. Cross-verification and deliberation guide revision and delivery, while construction tools, probe cases, and decision rules support subsequent tasks. We evaluate EnRealWorld on benchmark and open-ended world creation tasks and examine how experience transfers across tasks. Across models and tasks, EnRealWorld improves task completion, input fidelity, and physical responses to interventions, while accumulated experience transfers across tasks and agents, indicating its potential for scalable 4D world creation. The project page is available at \url{https://zihanguo1024.github.io/EnRealWorld/}.

\end{abstract}

\section{Introduction}
\label{sec:intro}

Constructing interactive virtual worlds from real-world observations and task descriptions provides controllable and reproducible testbeds for physical AI~\cite{zhang2026physical}.
To this end, recent advances in video world models~\cite{bruce2024genie} and explicit scene representations such as 3D Gaussian Splatting~\cite{kerbl20233dgs} have enabled dynamic generation and reconstruction~\cite{wu20244d,xie2024physgaussian}.
However, visual fidelity can coexist with incorrect physical behavior~\cite{motamed2026physicsiq,tang2026video2world, bansal2025videophy}. 
Overcoming this limitation requires moving toward executable worlds exposing physical states and mechanisms that can be inspected, modified, and tested under new actions~\cite{wang2026codeasworlds}.
Beyond generating interaction data with ground-truth state annotations, such worlds support diverse evaluation across long-horizon and out-of-distribution scenarios~\cite{wang2024gensim, wang2024robogen}. 
We formalize this task as \emph{4D world creation}: given multimodal inputs and task requirements, the goal is to construct an executable 4D world that explicitly represents physical entities, supports sustained interaction in a physics simulator, and remains editable and extensible.

\begin{figure*}[t]
    \centering
    \includegraphics[width=\textwidth]{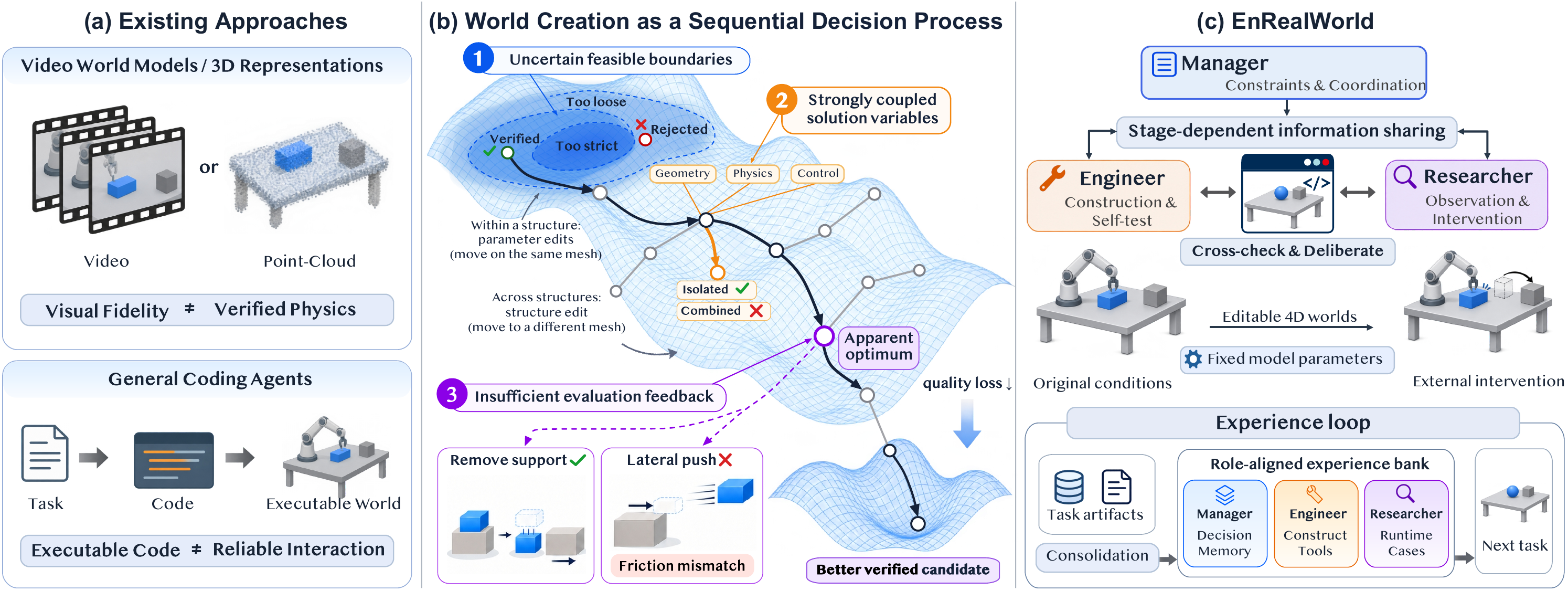}
    \caption{
    \textbf{Motivation and overview of EnRealWorld.}
    (a) Visual fidelity and executable code do not ensure reliable physical interaction.
    (b) World creation involves uncertain feasibility boundaries, coupled variables, and incomplete feedback.
    The landscape shows a possible search trajectory.
    (c) EnRealWorld combines complementary roles, stage-dependent information sharing, and cross-task experience reuse.
    }
    \label{fig:overview}
\end{figure*}

Manual world authoring requires specialist knowledge and extensive debugging.
While coding agents offer a promising avenue for automating this process at scale, directly applying general-purpose agents often fails to yield physically grounded environments~\cite{shao2026physcodegym}.
Without dedicated physical verification, generated worlds may bypass physical mechanisms through static frames, scripted animations, or hard-coded trajectories, appearing correct only under demonstration conditions.
From an optimization perspective, as shown in \cref{fig:overview}, 4D world creation is a feedback-driven search over a high-dimensional space of discrete and continuous variables. 
Because requirements and design consequences are only partly known in advance, solving this problem requires agents to alternate between clarifying the specification, revising the world, and running tests.

Three interdependent challenges arise.
\textbf{1) Uncertain feasibility boundaries.} 
Incomplete or ambiguous multimodal inputs and high-level intentions admit multiple scene structures and physical interpretations.
Overly restrictive constraints exclude valid implementations, and insufficient constraints admit incorrect worlds.
\textbf{2) Strongly coupled solution variables.}
Geometry, appearance, physical properties, and control logic are interdependent. 
Optimizations for isolated or familiar behaviors may fail under multi-body contact, composed operations, or long-horizon execution, hindering reliable assembly of locally effective components.
\textbf{3) Insufficient evaluation feedback.} 
Success under limited viewpoints and prescribed actions can conceal defective mechanisms. 
Failures requiring specific observations or interventions may remain undetected, trapping the search in superficially plausible pseudo-optimal regions.

To address these challenges, we introduce \textbf{EnRealWorld}, which formulates 4D world creation as agentic engineering via sequential decisions by complementary agents. 
The Manager translates multimodal inputs into implementation-independent propositions that define and refine the feasible region. 
The Engineer jointly constructs geometry, physics, and control, using discriminative experiments to diagnose coupled failures. 
The Researcher independently designs observations and physical probes to uncover failures the Engineer's self-tests miss.
Across rounds, independent verification and team deliberation coordinate repair and delivery, while consolidated role-specific experience accumulates to guide subsequent tasks. 
Our main contributions are as follows:

\begin{enumerate}
    \item \textbf{Problem formulation.}
    We formulate 4D world creation as a sequential decision process coupling specification refinement, world construction, proactive investigation, and cross-task experience accumulation.
    \item \textbf{The EnRealWorld framework.}
    We coordinate complementary agents through stage-dependent information boundaries and deliberation, with role-specific experience supporting reuse across tasks.
   \item \textbf{Systematic evaluation.}
    Benchmark and open-ended experiments across models show performance gains in task completion, input fidelity, and physical responses, alongside experience transfer across tasks and agents.
\end{enumerate}

\section{Formulation}
\label{sec:formulation}

\subsection{Definition and Goal of 4D World Creation}
\label{sec:formulation-definition}

\paragraph{Task Inputs and World Representation.}
The input is $x=(u,\mathcal O,\mathcal A)$, comprising natural-language requirements, observations, and available assets with their attributes. 
Any of these components may be absent.
An executable world $W=(\mathcal G,\mathcal P,\mathcal I)$ specifies entities, geometry, and appearance ($\mathcal G$), physical models, constraints, and parameters ($\mathcal P$), and control logic and interaction interfaces ($\mathcal I$).
The world is implemented as executable code with dependent assets.

\paragraph{Dynamics and Interaction.}
A backend $b$ executes the world through $s_{\tau+1}=f_{W,b}(s_\tau,a_\tau;\eta)$ and $o_\tau=g_{W,b}(s_\tau)$, 
where $\tau$ indexes simulation steps, 
$s_\tau$ is the queryable state, 
$a_\tau$ is an action or intervention, 
$\eta$ is the runtime configuration, 
and $o_\tau$ is a rendered or sensor observation.
The world must produce physically valid responses to specified interactions beyond replaying a reference trajectory.

\subsection{Sequential Decision Process}
\label{sec:formulation-sdp}

\paragraph{Construction State.}
We formulate world creation as a sequential decision process~\cite{puterman2014markov}.
At step $k$, the state $z_k=(\mathcal C_k,W_k,\mathcal F_k)$ comprises requirements, assumptions, and unresolved questions ($\mathcal C_k$), the candidate world ($W_k$), and accumulated observations and test outcomes ($\mathcal F_k$).

\begin{figure*}[t]
    \centering
    \includegraphics[width=\textwidth]{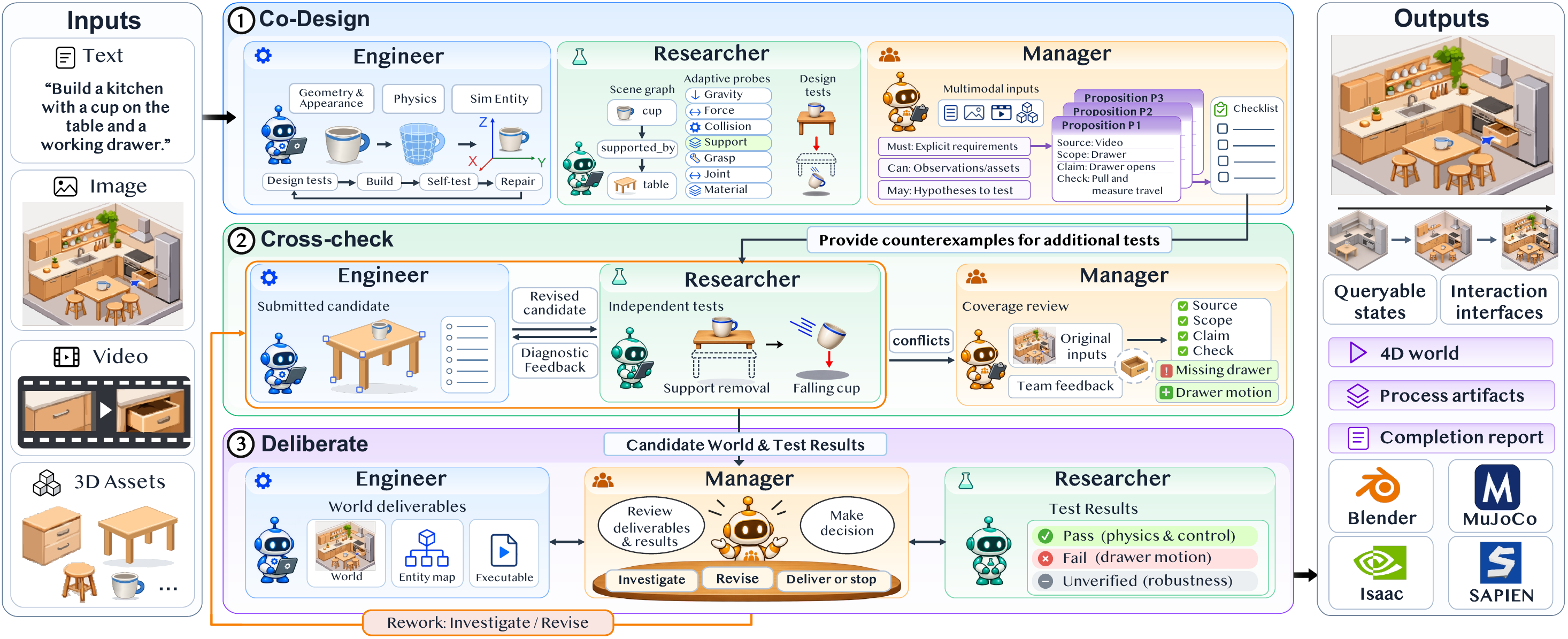}
    \caption{
    \textbf{Single-task workflow of EnRealWorld.}
    Co-Design combines specification, construction, and independent probe planning.
    Cross-check exchanges test results and guides revision.
    Deliberate determines whether to continue or conclude construction.
    }
    \label{fig:workflow}
\end{figure*}

\paragraph{Actions and State Transitions.}
Conditioned on input $x$, state $z_k$, and persistent experience $\mathcal M$, the policy selects an operation $d_k\sim\pi(\cdot\mid x,z_k,\mathcal M)$.
The transition kernel $\mathcal T$ determines the next construction state 
$z_{k+1}\sim\mathcal T(\cdot\mid z_k,d_k,x)$.
Specification refinement updates $\mathcal C_k$,
construction and revision update $W_k$,
and observations and interventions expand $\mathcal F_k$.
Delivery terminates the process.

\begin{table*}[t]
\centering
\caption{\textbf{Stage-dependent information visibility $\mathcal{V}_r(z_k)$ across roles and workflow stages.} $\checkmark$ marks role-scoped read/write or full read access, $\circ$ marks read-only access to frozen or accepted snapshots, and $\times$ marks hidden artifacts.}
\label{tab:supp_visibility}
\small
\setlength{\tabcolsep}{4pt}
\begin{tabular}{llccc}
\toprule
Stage & State Artifact & Manager ($\mathrm{M}$) & Engineer ($\mathrm{E}$) & Researcher ($\mathrm{R}$) \\
\midrule
\multirow{3}{*}{1. Co-Design} & Original input $x$ & $\checkmark$ & $\checkmark$ & $\checkmark$ \\
 & Draft specification $\mathcal{C}_k$ & $\checkmark$ (author) & $\times$ (until accepted) & $\times$ \\
 & Interim world $W_k$ \& self-tests & $\times$ & $\checkmark$ (author) & $\times$ \\
\midrule
\multirow{3}{*}{2. Cross-check} & Accepted specification $\mathcal{C}_k$ & $\checkmark$ (update) & $\circ$ & $\circ$ \\
 & Frozen candidate $W_k$ \& entity map & $\circ$ (audit) & $\checkmark$ (repair copy) & $\circ$ (read-only probe) \\
 & Probe findings \& repair traces $\mathcal{F}_k$ & $\circ$ & $\circ$ / $\checkmark$ (self-test) & $\checkmark$ (probe report) \\
\midrule
\multirow{3}{*}{3. Deliberate} & Updated specification $\mathcal{C}_k$ & $\checkmark$ & $\circ$ (accepted) & $\times$ \\
 & Reviewed candidate $W_k$ & $\circ$ & $\circ$ & $\circ$ \\
 & Aggregated test outcomes $\mathcal{F}_k$ \& open issues & $\checkmark$ (decide) & $\circ$ (respond) & $\circ$ (respond) \\
\bottomrule
\end{tabular}
\end{table*}

\subsection{Search Space and Objective}
\label{sec:formulation-space}

\paragraph{Search Space.}
The implementation space $\Omega$ combines discrete structural choices with continuous geometric and physical parameters.
$\Omega_{\mathrm{valid}}\subseteq\Omega$ denotes worlds satisfying mandatory requirements $\mathcal C_{\mathrm{hard}}$.
The specification $\mathcal C_k$ clarifies these requirements and their interpretation.

\paragraph{Optimization Objective.}
The ideal objective minimizes the loss $\mathcal L(W;x,\mathcal D_{\mathrm{eval}})$ over $\Omega_{\mathrm{valid}}$, where $\mathcal D_{\mathrm{eval}}$ specifies evaluation conditions.
The loss captures input fidelity, physical consistency, and interaction robustness.
During construction, executable checks assess these properties without optimizing a scalar loss.
Physically correct intervention responses may differ from reference trajectories.

\paragraph{Validation and Termination.}
Requirements are evaluated using $c_i=(\mathcal D_i,\phi_i)$, where $\mathcal D_i$ specifies initial states, actions, or interventions, and $\phi_i$ evaluates trajectories, observations, or simulator states.
Outcomes are \emph{Pass}, \emph{Fail}, or \emph{Unverified}.
Affected behaviors are retested after revision.
Construction ends when requirements are satisfied, or the budget is exhausted.
Task $n$ returns the final world, process artifacts $\mathcal B_n$, test outcomes, and unresolved issues.

\subsection{Experience and Cross-Task Evolution}
\label{sec:formulation-experience}

\paragraph{Experience-Conditioned Execution.}
Let $\mathcal M_n$ denote the experience available for task $n$.
It conditions the construction policy while the base model parameters $\theta$ remain fixed.

\paragraph{Cross-Task Experience Update.}
After task $n$, artifacts update experience through $\mathcal M_{n+1}=\mathcal U(\mathcal M_n,\mathcal B_n)$~\cite{zhao2024expel}.
Updates retain, extend, revise, or discard entries using internal observations, execution outcomes, and test records, without external benchmark scores.
The updated experience guides both construction and investigation in subsequent tasks.
Expanded definitions, two-timescale dynamics, and the quality objective appear in \cref{sec:supp_formulation}.

\section{Methodology}
\label{sec:method}

EnRealWorld controls information access across roles, enabling initial checks independent of builder assumptions.
\cref{fig:workflow} summarizes the workflow. Concrete SMT queries and runtime role prompts appear in \cref{sec:supp_method}.

\subsection{Overview}
\label{sec:method-overview}

\paragraph{Role Specialization.}
We define three role-specific construction policies as
$d_k^r\sim\pi_r\bigl(\cdot\mid x,\mathcal V_r(z_k),\mathcal M^r\bigr)$,
where $r\in\{\mathrm M,\mathrm E,\mathrm R\}$, Manager ($\mathrm M$) maintains $\mathcal C_k$, Engineer ($\mathrm E$) constructs and revises $W_k$, and Researcher ($\mathrm R$) acquires diagnostic results for $\mathcal F_k$.
Each role uses its experience $\mathcal M^r$ and a stage-dependent view $\mathcal V_r(z_k)$ of the construction state (\cref{tab:supp_visibility}).

\paragraph{Independent Inspection.}
Motivated by the limitations of self-correction~\cite{huang2024large}, we use separate sessions and asymmetric information boundaries to support independent checks.
For example, the Researcher derives verification criteria directly from the original input, independently of the Engineer's interpretation. 
The Engineer's interim results remain hidden from other roles during independent preparation. 
These probes complement the Engineer's self-tests and assess the submitted candidate during cross-checking.

\paragraph{Collaborative Refinement.}
Each round comprises:
\textbf{1) Co-Design.}
From the original input, the Manager develops propositions, the Engineer builds and self-tests the world, and the Researcher independently plans probes, in parallel.
\textbf{2) Cross-check.}
The Researcher probes the candidate, the Engineer repairs failures, and affected behaviors are retested.
The Manager updates the specification using input and team feedback.
\textbf{3) Deliberate.}
The Manager leads a review of requirement coverage and unresolved issues to choose clarification, investigation, revision, or delivery.

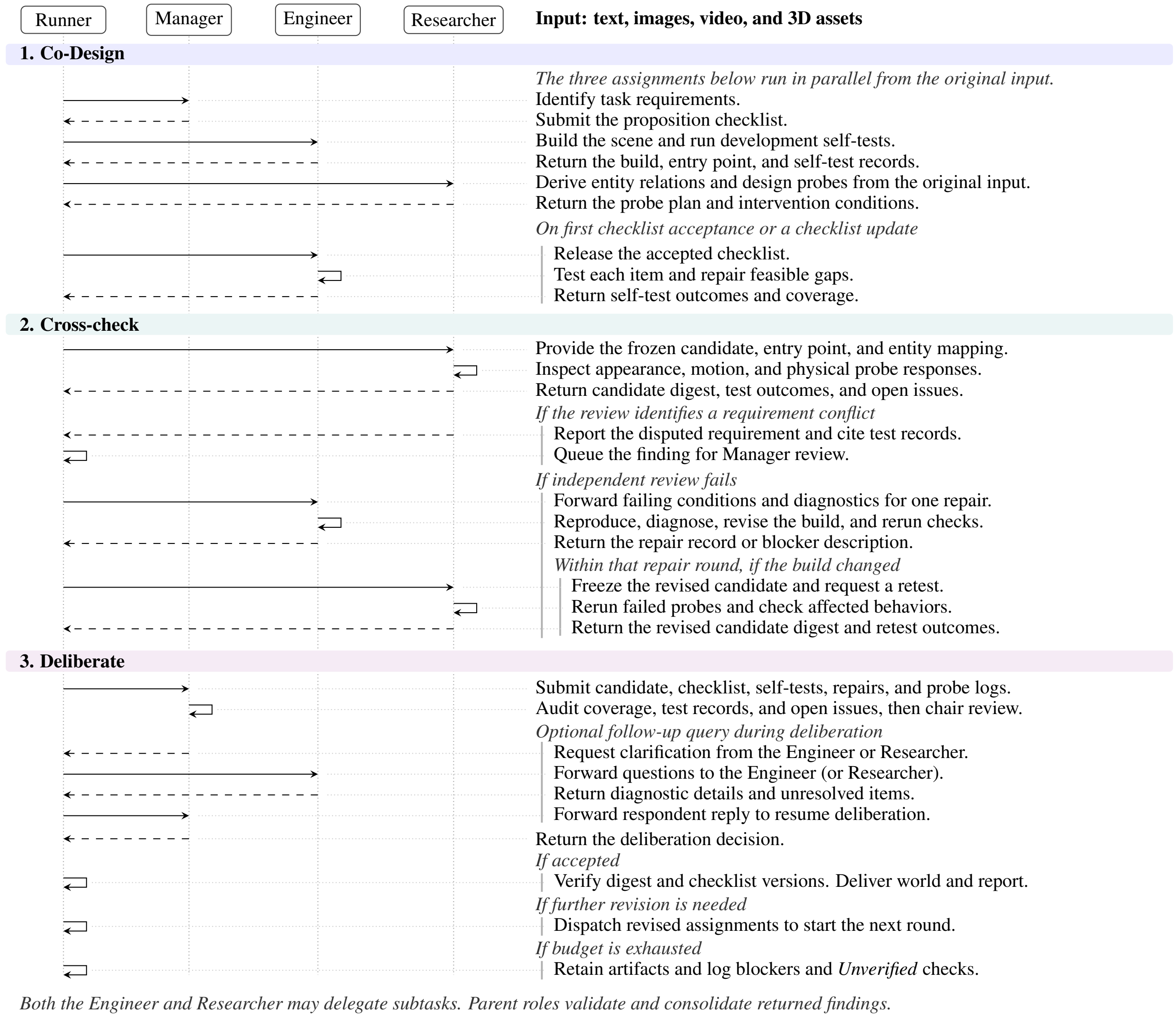
\begin{figure*}[t]
\centering
\begin{tikzpicture}[x=1cm,y=-0.34cm,font=\footnotesize,
  role/.style={draw=black!75,rounded corners=2pt,fill=white,inner sep=3pt,minimum width=1.25cm},
  action/.style={anchor=west,inner sep=0pt},
  guide/.style={densely dotted,gray!40}]
\def\stage#1#2#3{\fill[#3,rounded corners=1.5pt] (-0.85,#1-0.46) rectangle (16.35,#1+0.46);
  \node[action,font=\footnotesize\bfseries] at (-0.65,#1) {#2};}
\def\cond#1#2#3{\node[action,font=\footnotesize\itshape,text=black!80] at (#2,#1) {#3};}
\def\scopebar#1#2#3{\draw[gray!60,line width=0.8pt] (#1,#2) -- (#1,#3);}
\def\msg#1#2#3#4#5#6{\draw[->,>=stealth,#4] (#2,#1) -- (#3,#1);
  \draw[guide] ({max(#2,#3)+0.14},#1) -- (#5-0.12,#1);
  \node[action] at (#5,#1) {#6};}
\def\lstep#1#2#3#4{\draw[->,>=stealth] (#2,#1-0.20) -- ++(0.34,0)
  -- ++(0,0.40) -- (#2,#1+0.20);
  \draw[guide] (#2+0.42,#1) -- (#3-0.12,#1);
  \node[action] at (#3,#1) {#4};}

\foreach \x/\name in {0/Runner,1.85/Manager,3.75/Engineer,5.75/Researcher}{
  \draw[densely dotted,gray!70] (\x,0) -- (\x,41.4);
  \node[role] at (\x,0) {\name};}
\node[action,font=\footnotesize\bfseries] at (6.95,0) {Input: text, images, video, and 3D assets};

\stage{1.5}{1. Co-Design}{blue!8}
\cond{2.6}{6.95}{The three assignments below run in parallel from the original input.}
\msg{3.5}{0}{1.85}{solid}{6.95}{Identify task requirements.}
\msg{4.4}{1.85}{0}{dashed}{6.95}{Submit the proposition checklist.}
\msg{5.3}{0}{3.75}{solid}{6.95}{Build the scene and run development self-tests.}
\msg{6.2}{3.75}{0}{dashed}{6.95}{Return the build, entry point, and self-test records.}
\msg{7.1}{0}{5.75}{solid}{6.95}{Derive entity relations and design probes from the original input.}
\msg{8.0}{5.75}{0}{dashed}{6.95}{Return the probe plan and intervention conditions.}
\cond{9.1}{6.95}{On first checklist acceptance or a checklist update}
\scopebar{7.05}{9.8}{12.3}
\msg{10.2}{0}{3.75}{solid}{7.22}{Release the accepted checklist.}
\lstep{11.1}{3.75}{7.22}{Test each item and repair feasible gaps.}
\msg{12.0}{3.75}{0}{dashed}{7.22}{Return self-test outcomes and coverage.}

\stage{13.2}{2. Cross-check}{teal!8}
\msg{14.3}{0}{5.75}{solid}{6.95}{Provide the frozen candidate, entry point, and entity mapping.}
\lstep{15.2}{5.75}{6.95}{Inspect appearance, motion, and physical probe responses.}
\msg{16.1}{5.75}{0}{dashed}{6.95}{Return candidate digest, test outcomes, and open issues.}
\cond{17.1}{6.95}{If the review identifies a requirement conflict}
\scopebar{7.05}{17.6}{19.3}
\msg{18.0}{5.75}{0}{dashed}{7.22}{Report the disputed requirement and cite test records.}
\lstep{18.9}{0}{7.22}{Queue the finding for Manager review.}
\cond{20.0}{6.95}{If independent review fails}
\scopebar{7.05}{20.5}{26.8}
\msg{20.9}{0}{3.75}{solid}{7.22}{Forward failing conditions and diagnostics for one repair.}
\lstep{21.8}{3.75}{7.22}{Reproduce, diagnose, revise the build, and rerun checks.}
\msg{22.7}{3.75}{0}{dashed}{7.22}{Return the repair record or blocker description.}
\cond{23.7}{7.22}{Within that repair round, if the build changed}
\scopebar{7.32}{24.2}{26.7}
\msg{24.6}{0}{5.75}{solid}{7.48}{Freeze the revised candidate and request a retest.}
\lstep{25.5}{5.75}{7.48}{Rerun failed probes and check affected behaviors.}
\msg{26.4}{5.75}{0}{dashed}{7.48}{Return the revised candidate digest and retest outcomes.}

\stage{27.8}{3. Deliberate}{violet!8}
\msg{29.0}{0}{1.85}{solid}{6.95}{Submit candidate, checklist, self-tests, repairs, and probe logs.}
\lstep{29.9}{1.85}{6.95}{Audit coverage, test records, and open issues, then chair review.}
\cond{30.9}{6.95}{Optional follow-up query during deliberation}
\scopebar{7.05}{31.3}{34.8}
\msg{31.8}{1.85}{0}{dashed}{7.22}{Request clarification from the Engineer or Researcher.}
\msg{32.7}{0}{3.75}{solid}{7.22}{Forward questions to the Engineer (or Researcher).}
\msg{33.6}{3.75}{0}{dashed}{7.22}{Return diagnostic details and unresolved items.}
\msg{34.5}{0}{1.85}{solid}{7.22}{Forward respondent reply to resume deliberation.}
\msg{35.5}{1.85}{0}{dashed}{6.95}{Return the deliberation decision.}
\cond{36.5}{6.95}{If accepted}
\scopebar{7.05}{37.0}{37.8}
\lstep{37.4}{0}{7.22}{Verify digest and checklist versions. Deliver world and report.}
\cond{38.4}{6.95}{If further revision is needed}
\scopebar{7.05}{38.9}{39.7}
\lstep{39.3}{0}{7.22}{Dispatch revised assignments to start the next round.}
\cond{40.3}{6.95}{If budget is exhausted}
\scopebar{7.05}{40.8}{41.6}
\lstep{41.2}{0}{7.22}{Retain artifacts and log blockers and \emph{Unverified} checks.}

\node[action,font=\footnotesize\itshape,text=black!80] at (-0.65,42.7)
  {Both the Engineer and Researcher may delegate subtasks. Parent roles validate and consolidate returned findings.};
\end{tikzpicture}
\caption{\textbf{Scheduling within one construction round.} Solid arrows show runner dispatches and handoffs, dashed arrows return reports or test outcomes, loops mark single-role steps, and vertical bars group conditional branches.}
\label{fig:supp_role_sequence}
\end{figure*}

Across stages, both the Engineer and Researcher may delegate separable subtasks to sub-agents, while parent roles validate returned findings against the task and candidate digest.
Process completion and behavioral verification are recorded separately, so a clean self-test run does not certify obstacle response unless contact forces or blocking are observed.
\cref{fig:supp_role_sequence} details the per-round runner handoffs, and \cref{alg:enrealworld} summarizes the end-to-end refinement and experience-update loop.

\begin{algorithm}[tb]
\caption{Single-Task 4D World Construction and Role-Specific Experience Update}
\label{alg:enrealworld}
\small
\renewcommand{\algorithmicrequire}{\textbf{Input:}}
\renewcommand{\algorithmicensure}{\textbf{Output:}}
\begin{algorithmic}[1]
\raggedright
\Require Task $x_n=(u,\mathcal{O},\mathcal{A})$, experience $\mathcal{M}_n=(\mathcal{M}_n^r)$, budget
\Ensure Deliverables $\mathcal{Y}_n=(W_n^{\mathrm{out}},\mathcal{B}_n,\mathcal{R}_n)$, updated $\mathcal{M}_{n+1}$
\State Initialize $z_0=(\mathcal{C}_0,W_0,\mathcal{F}_0)$
\While{budget remains}
    \State Let $k$ index the construction state at the start of this round
    \Statex \vspace{2pt}\hspace{\algorithmicindent}\textcolor{black!65}{\textit{// Stage 1: Co-Design (parallel, stage view $\mathcal{V}_r$)}}
    \State $\mathrm{M}$: draft/refine $\mathcal{C}_k$ from $x_n,\mathcal{M}_n^{\mathrm{M}}$ (Z3 for formalizable conditions)
    \State $\mathrm{E}$: build $W_k$ and run self-tests $\mathcal{T}_{\mathrm{E}}$ via $\mathcal{M}_n^{\mathrm{E}}$
    \State $\mathrm{R}$: derive scene graph $\mathcal{S}$ and probe plan via $\mathcal{M}_n^{\mathrm{R}}$
    \Statex \vspace{2pt}\hspace{\algorithmicindent}\textcolor{black!65}{\textit{// Stage 2: Cross-check (frozen by digest)}}
    \State Runner: freeze $W_k$ for review with entry points and entity map $\mathcal{K}$
    \State $\mathrm{R}$: run probes $q=(\psi,a,o,\phi)$ on $W_k$, log to $\mathcal{F}_k$
    \State $\mathrm{R}$: report requirement conflicts and findings to $\mathrm{M}$
    \State $\mathrm{E}$: repair affected units $\Delta W$ on failure
    \State $\mathrm{M}$: audit findings and refine $\mathcal{C}_{k+1}=\Gamma(\mathcal{C}_k;x_n,\mathcal{F}_k)$
    \State On a changed digest, freeze the revision for $\mathrm{R}$ to retest affected findings
    \Statex \vspace{2pt}\hspace{\algorithmicindent}\textcolor{black!65}{\textit{// Stage 3: Deliberate}}
    \State $\mathrm{M}$: audit checklist coverage against self-tests, repairs, and independent retests in $\mathcal{F}_k$
    \While{clarification is needed and budget remains}
        \State $\mathrm{M}$: question $\mathrm{E}$ or $\mathrm{R}$ about an open issue. The queried role reports its test records and any remaining uncertainty.
    \EndWhile
    \If{an open issue admits a distinguishing test within budget}
        \State $\mathrm{M}$: continue with next-round assignments for investigation or revision
    \Else
        \State $\mathrm{M}$: request delivery if required checks passed, or report failure with blockers and unmet requirements
        \State Runner: accept a pass only if the reviewed digest matches the candidate and the current checklist version was handed to $\mathrm{E}$ for self-testing
        \State \textbf{if} the runner accepts the finish decision \textbf{then break}
    \EndIf
\EndWhile
\State Assemble deliverables $\mathcal{Y}_n=(W_n^{\mathrm{out}},\mathcal{B}_n,\mathcal{R}_n)$
\ForAll{$r \in \{\mathrm{M},\mathrm{E},\mathrm{R}\}$}
    \State Extract $\mathcal{B}_n^r \subseteq \mathcal{B}_n$, update $\mathcal{M}_{n+1}^r \gets \mathcal{U}^r(\mathcal{M}_n^r,\mathcal{B}_n^r)$
\EndFor
\State \Return $\mathcal{Y}_n$ and $\mathcal{M}_{n+1}$
\end{algorithmic}
\end{algorithm}

\subsection{4D World Specification via Propositions}
\label{sec:method-proposition}

\paragraph{Proposition Specification.}
The Manager maintains a checklist of implementation-independent propositions, each specifying its source, scope, claim, and check.
Explicit requirements (\emph{Must}) define acceptance criteria, supported inferences (\emph{Can}) guide construction, and unresolved hypotheses (\emph{May}) guide observations and interventions.
The Manager merges overlapping information, flags conflicts, and recovers omitted requirements across modalities.
Explicit requirements take precedence over inferences and hypotheses, while unresolved conflicts guide further investigation.

\paragraph{Consistency Checking and Refinement.}
To ensure logical soundness, the Manager employs Z3~\cite{de2008z3}, a satisfiability modulo theories (SMT) solver, to check formalizable requirements and assumptions for consistency.
Conflicts are returned for clarification before construction continues. Concrete Manager and Researcher Z3 queries appear in \cref{sec:supp_b2}.
Test outcomes guide specification updates $\mathcal{C}_{k+1}=\Gamma(\mathcal{C}_k;x,\mathcal{F}_k)$ to address omissions, revise assumptions, and resolve hypotheses.
Updates remain grounded in the input and preserve explicit requirements, preventing success through relaxed acceptance criteria.

\subsection{Iterative World Construction with Self-Tests}
\label{sec:method-construction}

\paragraph{Behavioral Self-Tests.}
Using the task input and available team feedback, the Engineer constructs the candidate world across geometry, physics, and control, and derives behavioral self-tests $\mathcal T_{\mathrm E}=\{(\mathcal D_i,\phi_i)\}_i$.
Tests cover nominal behavior, boundary conditions, and expected failures, using measurable quantities such as hold duration, contact impulse, penetration depth, and drift.
Checks use simulator states for physics and rendered views for appearance.

\paragraph{Diagnosing Failed Self-Tests.}
When a self-test fails, the Engineer designs experiments to distinguish geometric, physical, and control-related causes.
For example, disabling control while holding initial states and collision geometry fixed distinguishes controller pose overrides from static body declarations or supporting contacts.
Each repair binds the failing condition and prior candidate digest to the revised world, reproducing the triggering failure first and rechecking neighboring behaviors for regressions.
Non-physical edits explicitly reference prior simulation rollouts so stale trajectories are never treated as newly verified physics.
The Engineer submits the candidate world with execution entry points and entity mappings for the Researcher's independent probes.

\subsection{Proactive Investigation with Physical Probes}
\label{sec:method-intervention}

\paragraph{Probing with Scene Graphs.}
The Researcher selects observations and interventions to reveal hidden failures or distinguish their causes.
It independently derives a scene graph from the input and maps its entities to the candidate world.
Physical relations such as support, contact, and articulation guide probe selection and failure localization.
The Researcher prioritizes probes using scene relations, unresolved requirements, revisions, and observed anomalies.
Follow-up probes distinguish competing explanations of delivery-relevant failures.
For a formalizable assertion $\psi$, let $\mathcal{C}_{k,\psi}^{\mathrm{bg}}$ be the conjunction of independently specified, jointly satisfiable background constraints excluding $\psi$.
The Researcher queries the solver for a counterexample:
\begin{equation}
\operatorname{SAT}\!\left(\mathcal{C}_{k,\psi}^{\mathrm{bg}} \land \neg\psi\right).
\end{equation}

A satisfying assignment suggests a simulation state or intervention to test, whereas \texttt{unsat} establishes entailment within the encoding.

\paragraph{Counterfactual Probes.}

A probe $q=(\psi,a,o,\phi)$ specifies an assertion, intervention, observation, and response check.
The Researcher compares intervened and reference executions, e.g., removing support tests whether an object falls, and changing force direction tests whether a joint constraint remains effective.
Our probes cover seven physical failure families summarized in \cref{tab:supp_probespec}, from gravity and contact mechanics to joint constraints and material response.
Let $\mathbf{s}_{0:T}(W,a)$ denote the simulated state trajectory under intervention $a$, with $a=\varnothing$ denoting the baseline retaining original task actions.
Intervention sensitivity for probe $q$ is measured by the trajectory deviation
\begin{equation}
\delta_q = \operatorname{dist}\!\left(\mathbf{s}_{0:T}(W,a),\,\mathbf{s}_{0:T}(W,\varnothing)\right).
\end{equation}
Because deviation magnitude alone does not certify physical correctness, the predicate $\phi$ checks directional, contact, and kinematic-limit criteria, such as verifying downward motion after withdrawing a supporting surface.

\paragraph{Controlled Testing and Targeted Retesting.}
The Researcher controls other conditions and evaluates intervention responses using simulator states and aligned rendered views.
Failure reports identify conditions, affected entities, and expected responses.
Revisions trigger entity remapping and retesting of affected behaviors.
Changes to shared parameters or coupled interactions trigger broader checks.

\begin{table*}[t]
\centering
\caption{\textbf{Representative paired interventions and diagnostic readouts across the seven physical probe families.} Each pair holds the candidate world and all non-intervened conditions fixed.}
\label{tab:supp_probespec}
\small
\begin{tabular}{@{}p{.16\textwidth}p{.34\textwidth}p{.45\textwidth}@{}}
\toprule
Probe Family & Paired Conditions & Diagnostic Readout for Predicate $\phi$ \\
\midrule
Gravity \& free motion & Hold vs.\ release an unconstrained body & Post-release displacement and velocity trajectory \\
Force response & Perturb direction or magnitude of applied force & Induced acceleration direction and dynamic response \\
Collision & Retain vs.\ remove an obstacle along the path & Contact forces, blocking behavior, and penetration depth \\
Support & Retain vs.\ withdraw the sole supporting surface & Loss of support contact and downward acceleration/motion \\
Grasping & Maintain vs.\ release gripper actuation & Relative end-effector--object pose, slip, and release motion \\
Joint constraints & Apply loads along compliant vs.\ locked axes & Active motion axis, joint travel, and limit enforcement \\
Material behavior & Perturb friction, restitution, or applied load & Sliding distance, rebound height, or deformation profile \\
\bottomrule
\end{tabular}
\end{table*}

\begin{table*}[!t]
\centering
\caption{
\textbf{Video2World results across seven task families.}
Each cell reports the mean score $S$ above the mean function, geometry, and dynamics scores ($F \mid G \mid D$).
Within each model, the highest $S$ in each column is bolded and the second highest is underlined.
}
\label{tab:video2world_benchmark}
\footnotesize
\setlength{\tabcolsep}{2pt}

\definecolor{v2wModel}{RGB}{222,215,234}
\definecolor{v2wBase}{RGB}{232,232,232}
\definecolor{v2wBaselines}{RGB}{232,232,232}
\definecolor{v2wOurs}{RGB}{228,239,247}

\newcommand{\vscore}[2]{%
\ensuremath{\vcenter{\hbox{\shortstack[c]{#1\\$#2$}}}}%
}
\newcommand{\vfirst}[1]{\textbf{#1}}
\newcommand{\vsecond}[1]{\underline{#1}}
\newcommand{\vna}{\mbox{--}}

\newcommand{\vmodel}[1]{%
  \rowcolor{v2wModel}\multicolumn{9}{c}{\textbf{#1}}\\%
}
\newcommand{\vgroup}[2]{%
  \hline
  \rowcolor{#1}\multicolumn{9}{l}{\textbf{#2}}\\%
}

\begin{tabular*}{\textwidth}{@{\extracolsep{\fill}}l|c|ccccccc@{}}
\toprule
\multicolumn{1}{c|}{Method} & Overall & DexYCB & DROID & HOI4D & HOT3D
       & In-house & OakInk & RoboDojo \\
\midrule
\vmodel{GPT-6.1 Sol}

\vgroup{v2wBase}{Base Agent}
Base Agent
 & \vscore{46.00}{26.8|64.5|46.7}
 & \vscore{62.62}{58.4|63.3|66.1}
 & \vscore{50.41}{39.9|65.8|45.5}
 & \vscore{28.23}{12.5|42.5|29.7}
 & \vscore{\vsecond{40.20}}{12.5|84.5|23.7}
 & \vscore{40.26}{18.3|60.5|41.9}
 & \vscore{\vfirst{54.08}}{18.8|73.9|69.5}
 & \vscore{57.09}{3.8|78.8|88.7} \\

\vgroup{v2wBaselines}{Baselines}
VIGA
 & \vscore{\vsecond{51.21}}{40.8|66.5|46.3}
 & \vscore{\vsecond{68.73}}{58.4|81.6|66.3}
 & \vscore{\vsecond{58.38}}{56.5|70.5|48.1}
 & \vscore{\vsecond{33.86}}{25.0|46.6|30.0}
 & \vscore{35.31}{6.2|77.0|22.6}
 & \vscore{44.30}{32.2|59.1|41.6}
 & \vscore{47.27}{18.8|67.5|55.6}
 & \vscore{72.56}{50.0|87.7|80.0} \\
SceneSmith
 & \vscore{47.36}{34.7|62.4|44.9}
 & \vscore{27.12}{8.3|49.9|23.1}
 & \vscore{\vfirst{59.32}}{56.0|68.5|53.5}
 & \vscore{33.01}{25.0|43.8|30.2}
 & \vscore{32.64}{6.2|69.3|22.4}
 & \vscore{40.27}{25.3|58.8|36.7}
 & \vscore{40.52}{6.2|50.3|65.0}
 & \vscore{\vsecond{72.85}}{50.0|83.3|85.2} \\
SceneWeaver
 & \vscore{49.14}{33.3|65.5|48.6}
 & \vscore{45.17}{16.7|81.8|37.0}
 & \vscore{54.08}{43.2|67.7|51.4}
 & \vscore{\vfirst{40.08}}{37.5|54.4|28.3}
 & \vscore{\vfirst{42.52}}{12.5|80.9|34.2}
 & \vscore{43.43}{27.5|58.7|44.1}
 & \vscore{51.40}{25.0|63.5|65.7}
 & \vscore{\vfirst{74.89}}{50.0|81.8|92.8} \\
Code2Worlds-2
 & \vscore{47.26}{29.0|67.1|45.7}
 & \vscore{57.13}{58.4|78.3|34.8}
 & \vscore{51.79}{41.4|69.5|44.5}
 & \vscore{25.00}{6.2|48.4|20.4}
 & \vscore{34.64}{6.2|75.0|22.7}
 & \vscore{\vsecond{44.80}}{25.3|63.5|45.6}
 & \vscore{48.11}{12.5|58.4|73.4}
 & \vscore{58.08}{3.8|84.6|85.9} \\

\vgroup{v2wOurs}{Ours}
EnRealWorld
 & \vscore{\vfirst{53.55}}{53.2|60.1|47.3}
 & \vscore{\vfirst{70.25}}{100.0|58.5|52.3}
 & \vscore{50.17}{38.4|65.8|46.3}
 & \vscore{28.40}{12.5|40.2|32.5}
 & \vscore{30.33}{12.5|69.8|8.6}
 & \vscore{\vfirst{58.88}}{77.5|52.1|47.1}
 & \vscore{\vsecond{51.80}}{12.5|76.4|66.5}
 & \vscore{70.71}{50.0|76.0|86.1} \\

\midrule
\vmodel{DeepSeek-V4.1-Flash-1017}

\vgroup{v2wBase}{Base Agent}
Base Agent
 & \vscore{39.43}{25.5|54.4|38.4}
 & \vscore{\vfirst{73.75}}{100.0|56.2|65.0}
 & \vscore{41.49}{31.2|56.7|36.5}
 & \vscore{\vsecond{25.31}}{18.8|42.8|14.4}
 & \vscore{\vfirst{28.76}}{18.8|60.3|7.2}
 & \vscore{34.17}{17.5|50.2|34.8}
 & \vscore{\vsecond{47.14}}{6.2|67.0|68.1}
 & \vscore{47.23}{3.8|61.4|76.5} \\

\vgroup{v2wBaselines}{Baselines}
VIGA
 & \vscore{35.98}{17.4|54.8|35.7}
 & \vscore{54.44}{58.4|61.3|43.8}
 & \vscore{39.79}{21.1|61.6|36.6}
 & \vscore{\vfirst{27.64}}{18.8|43.5|20.8}
 & \vscore{22.47}{0.0|56.5|10.9}
 & \vscore{28.65}{7.8|45.6|32.6}
 & \vscore{41.55}{6.2|64.2|54.1}
 & \vscore{62.07}{50.0|69.4|66.8} \\
SceneSmith
 & \vscore{38.55}{30.0|51.3|34.3}
 & \vscore{29.94}{0.0|39.8|50.0}
 & \vscore{43.83}{38.7|57.8|35.0}
 & \vscore{22.29}{0.0|55.9|11.1}
 & \vscore{24.55}{12.5|48.4|12.8}
 & \vscore{\vsecond{37.55}}{26.9|48.4|37.3}
 & \vscore{6.20}{0.0|18.6|0.0}
 & \vscore{\vsecond{80.28}}{100.0|70.1|70.8} \\
SceneWeaver
 & \vscore{\vsecond{39.71}}{24.7|55.8|38.5}
 & \vscore{40.22}{0.0|61.1|59.6}
 & \vscore{\vfirst{49.94}}{42.0|61.3|46.6}
 & \vscore{18.45}{0.0|44.4|10.9}
 & \vscore{15.05}{0.0|42.6|2.6}
 & \vscore{34.93}{23.3|49.9|31.6}
 & \vscore{41.58}{6.2|74.0|44.5}
 & \vscore{47.37}{7.5|63.4|71.2} \\
Code2Worlds-2
 & \vscore{38.62}{26.3|51.7|37.8}
 & \vscore{41.69}{0.0|59.2|65.8}
 & \vscore{43.37}{35.4|52.2|42.5}
 & \vscore{23.03}{0.0|44.0|25.1}
 & \vscore{\vsecond{26.74}}{12.5|53.8|13.9}
 & \vscore{33.63}{25.6|47.5|27.8}
 & \vscore{41.28}{6.2|64.4|53.1}
 & \vscore{64.43}{53.8|65.8|73.8} \\

\vgroup{v2wOurs}{Ours}
EnRealWorld
 & \vscore{\vfirst{53.43}}{55.7|57.9|46.7}
 & \vscore{\vsecond{55.98}}{50.0|60.2|57.8}
 & \vscore{\vsecond{48.50}}{37.8|61.5|46.3}
 & \vscore{22.10}{6.2|38.3|21.7}
 & \vscore{19.72}{0.0|52.9|6.3}
 & \vscore{\vfirst{63.23}}{87.8|55.2|46.7}
 & \vscore{\vfirst{48.27}}{6.2|61.3|77.2}
 & \vscore{\vfirst{82.13}}{100.0|73.1|73.3} \\
 \midrule
\vmodel{Kimi-3.1 Pro}

\vgroup{v2wBase}{Base Agent}
Base Agent
 & \vscore{\vsecond{37.98}}{18.2|61.8|33.9}
 & \vscore{37.80}{0.0|77.6|35.8}
 & \vscore{\vsecond{44.05}}{31.6|62.2|38.4}
 & \vscore{17.94}{0.0|53.8|0.0}
 & \vscore{19.70}{0.0|55.5|3.7}
 & \vscore{29.50}{1.1|59.7|27.7}
 & \vscore{\vfirst{52.19}}{25.0|66.0|65.6}
 & \vscore{83.29}{100.0|69.8|80.1} \\

\vgroup{v2wBaselines}{Baselines}
VIGA
 & \vscore{34.71}{14.4|60.2|29.5}
 & \vscore{\vfirst{42.96}}{0.0|74.6|54.4}
 & \vscore{41.98}{31.0|59.5|35.5}
 & \vscore{\vsecond{19.48}}{0.0|58.4|0.0}
 & \vscore{\vsecond{24.35}}{0.0|57.3|15.8}
 & \vscore{30.41}{8.6|57.9|24.7}
 & \vscore{21.94}{0.0|65.5|0.3}
 & \vscore{46.14}{0.0|68.0|70.5} \\
SceneSmith
 & \vscore{34.37}{11.8|56.8|34.5}
 & \vscore{31.18}{0.0|41.4|52.2}
 & \vscore{41.90}{28.3|59.3|38.2}
 & \vscore{12.78}{0.0|38.4|0.0}
 & \vscore{21.80}{0.0|64.5|1.0}
 & \vscore{29.21}{3.3|54.9|29.4}
 & \vscore{38.69}{0.0|62.6|53.5}
 & \vscore{53.43}{7.5|74.9|77.9} \\
SceneWeaver
 & \vscore{33.17}{11.0|56.8|31.8}
 & \vscore{39.19}{0.0|79.5|38.1}
 & \vscore{41.03}{25.6|57.0|40.5}
 & \vscore{15.11}{0.0|45.3|0.0}
 & \vscore{18.55}{0.0|55.7|0.0}
 & \vscore{28.98}{4.2|52.9|29.9}
 & \vscore{25.20}{0.0|64.3|11.3}
 & \vscore{44.27}{3.8|66.2|62.9} \\
Code2Worlds-2
 & \vscore{36.89}{20.6|59.1|30.9}
 & \vscore{\vsecond{39.77}}{16.7|61.8|40.9}
 & \vscore{40.05}{25.6|59.7|34.9}
 & \vscore{13.51}{0.0|40.6|0.0}
 & \vscore{21.50}{0.0|62.4|2.1}
 & \vscore{\vsecond{34.14}}{14.2|57.0|31.3}
 & \vscore{24.60}{0.0|73.9|0.0}
 & \vscore{\vsecond{83.49}}{100.0|69.4|81.1} \\

\vgroup{v2wOurs}{Ours}
EnRealWorld
 & \vscore{\vfirst{47.14}}{43.2|57.5|40.8}
 & \vscore{36.83}{8.4|65.4|36.8}
 & \vscore{\vfirst{44.18}}{42.6|52.8|37.2}
 & \vscore{\vfirst{45.39}}{37.5|67.8|30.9}
 & \vscore{\vfirst{32.48}}{18.8|78.4|0.4}
 & \vscore{\vfirst{48.77}}{48.9|56.1|41.3}
 & \vscore{\vsecond{44.69}}{12.5|52.9|68.8}
 & \vscore{\vfirst{84.76}}{100.0|66.0|88.4} \\
\bottomrule
\end{tabular*}
\end{table*}

\subsection{Experience Persistence and Evolution}
\label{sec:method-experience}

\paragraph{Role-Specific Experience.}
Each role retrieves and applies entries from its own persistent library to guide task decisions.
The Engineer retains reusable construction, self-testing, and repair tools.
The Researcher stores runtime cases with preconditions, interventions, expected responses, observations, and counterexamples.
The Manager retains decision memory for requirement clarification and acceptance, including applicability conditions.

\paragraph{Experience Update.}
After each task, each role $r\in\{\mathrm{M},\mathrm{E},\mathrm{R}\}$ extracts role-specific internal artifacts $\mathcal{B}_n^r \subseteq \mathcal{B}_n$ and updates its library through addition, revision, merging, or pruning:
\begin{equation}
\mathcal{M}_{n+1} = \big(\mathcal{U}^r(\mathcal{M}_n^r, \mathcal{B}_n^r)\big)_{r\in\{\mathrm{M},\mathrm{E},\mathrm{R}\}}.
\end{equation}
In-place revisions refine verification standards without increasing library entry counts.
Role-specific experience preserves information isolation across tasks, allowing construction and verification methods to develop separately and provide complementary feedback in later tasks.

\subsection{Verifiability and Runtime Scaling}
\label{sec:supp_b6}

\paragraph{Verifiability conditions.}
Verifying an intervention-based behavioral proposition requires three preconditions: \emph{entity grounding} to bind every referenced entity to a native simulator object, \emph{intervention executability} so the backend supports the prescribed action or perturbation, and \emph{diagnostic observability} so readouts distinguish satisfaction from violation.
Repeating an unchanged probe cannot recover information absent from its readout, as a fixed exterior camera generally cannot resolve intersections between occluded mesh surfaces.
Complementary mechanisms cover other checks, with SMT solvers testing symbolic consistency and aligned renderings verifying directly visible appearance properties.

\paragraph{Execution cost accounting.}
Across $N_{\mathrm{rd}}$ deliberation rounds, let $c_t$, $m_t$, and $q_t$ count role invocations, inter-role messages, and executed probes in round $t$.
Cumulative execution volume is given by $\sum_{t=1}^{N_{\mathrm{rd}}} c_t$, $\sum_{t=1}^{N_{\mathrm{rd}}} m_t$, and $\sum_{t=1}^{N_{\mathrm{rd}}} q_t$.
Independent checks can run concurrently to reduce wall-clock latency, while synchronization points such as deliberation reviews wait on their slowest upstream dependency.

\paragraph{Dependency-guided retesting.}
Let $\operatorname{Dep}(\psi)$ denote the set of scene entities, physical parameters, and contact pairs on which assertion $\psi$ depends.
After a revision to units $\Delta W$, the framework prioritizes retesting assertions with $\operatorname{Dep}(\psi)\cap\Delta W\neq\varnothing$.
Shared material properties or coupled multibody contacts enlarge $\operatorname{Dep}(\psi)$ and broaden regression coverage.

\section{Experiments}

We evaluate EnRealWorld on benchmark and open-ended world creation tasks, and examine experience accumulation and transfer across tasks and agents.

\subsection{Experimental Setup}

\paragraph{Tasks.}
Video2World~\cite{tang2026video2world} evaluates executable world construction from videos of physical processes.
We additionally evaluate on the 43 counterfactual tasks from PhysCodeGym~\cite{shao2026physcodegym}.
Open-ended tasks use natural-language requirements, paintings, and architectural drawings.
Agents select assets from NVIDIA SimReady Assets~\cite{nvidiaSimReadyAssets} and ChemLab-Assets~\cite{song2026chemlabassets}, without scene-specific asset preselection by the authors.

\paragraph{Models.}
Benchmark experiments use Codex with GPT-6.1 Sol, DeepSeek-V4.1-Flash-1017, and Kimi-3.1 Pro.
Open-ended experiments use Codex with GPT-6.1 Sol.
Frontier-model experiments additionally use Codex with GPT-6 Astra and GPT-6.2 Astra, Claude Code with Claude Fable 5.2, and Antigravity with Gemini 4 Neon.
Within EnRealWorld, all roles and the main session share the same agent framework and model, with medium reasoning effort.

\paragraph{Baselines.}
We compare against VIGA~\cite{yin2026viga}, SceneSmith~\cite{pfaff2026scenesmith}, Code2Worlds-2~\cite{zhang2026code2worlds2}, SceneWeaver~\cite{sceneweaver}, and direct execution by the base agent (``Base Agent'').
Within each setting, methods share the base agent and model.

\paragraph{Evaluation details.}
Evaluation is independent of construction.
We use official Video2World metrics and procedures, the official isolated PhysCodeGym verifier, and agent-as-judge evaluation with human inspection for open-ended tasks.
Depending on the task, backends include Blender 4.0.2 or 5.2.2 LTS, MuJoCo 3.11.0, SAPIEN 3.0.3 with ManiSkill 3.0.1, and Isaac Sim 5.1.
Unless otherwise specified, each task starts with an empty experience library and retains no experience across tasks.
Details are provided in \cref{sec:supp_setup}.

\subsection{Benchmark Performance}

\paragraph{Overall performance and cross-model adaptation.}
EnRealWorld achieves the highest overall score across all three models (\cref{tab:video2world_benchmark}), with larger gains over Base Agent under DeepSeek and Kimi-3.1 Pro.
All baselines improve over Base Agent with GPT-6.1 Sol, whereas most fail to do so with DeepSeek and none beat Base Agent overall with Kimi-3.1 Pro.
On Kimi-3.1 Pro, EnRealWorld ranks first overall and in five of seven families.
The benefits of the other construction frameworks are therefore more sensitive to the base model in these experiments.
Signed per-family score differences over Base Agent across all three models are reported in \cref{sec:supp_bench} (\cref{tab:supp_v2w}).

\paragraph{Function, geometry, and dynamics.}
On In-house tasks with DeepSeek, EnRealWorld increases the function score from 17.5 to 87.8 relative to Base Agent, while improving geometry and dynamics.
For these tasks, the largest component gain comes from realizing the target functions, showing that plausible scene structure and local dynamic behavior do not necessarily ensure task completion.

\paragraph{Qualitative trajectories.}
\cref{fig:supp_v2w_qual} presents representative Video2World reconstructions produced by EnRealWorld with GPT-6.1 Sol, pairing source video frames with simulated executions across key manipulation stages. 
Following the official Video2World protocol, supported scenes use simplified object geometry or task-specified proxies to focus on object motion and manipulation stages.
Across rigid-body, deformable, and articulated tasks, EnRealWorld reproduces the motion trajectories and stage-wise contact transitions in the source videos.

\paragraph{PhysCodeGym.}
\cref{tab:supp_physcodegym} compares Base Agent and EnRealWorld on all 43 counterfactual tasks of PhysCodeGym~\cite{shao2026physcodegym} under GPT-6.1 Sol.
EnRealWorld reduces candidate failures from 19 to 6 while raising the mean visual score from 0.336 to 0.491 and the mean physics score from 0.321 to 0.416.

\paragraph{Token usage and runtime.}
Each Video2World task ran on one NVIDIA GeForce RTX 4090 GPU, 16 CPU cores, and 128 GB of RAM.
On GPT-6.1 Sol, EnRealWorld reaches 53.55 against 51.21 for VIGA, adding about 25 minutes per task on average.
On DeepSeek-V4.1-Flash-1017 and Kimi-3.1 Pro, it outscores the strongest competing method by 13.72 and 9.16 points with about 75 and 76 additional minutes per task.
EnRealWorld on DeepSeek-V4.1-Flash-1017 reaches 53.43, surpassing every baseline on GPT-6.1 Sol.
Excluding one GPT-6.1 Sol run interrupted by an API error, cached reads account for $97.4\%$ to $98.6\%$ of EnRealWorld's input tokens across models, leaving $1.4\%$ to $2.6\%$ as uncached fresh input.

\subsection{Component Ablations}
\label{sec:supp_ablation}

\paragraph{Configurations.}
We evaluate five configurations under GPT-6.1 Sol in the same execution environment.
\textbf{Base Agent} executes the task directly in a single session.
\textbf{ERW-Prompts} adds the role responsibility prompts from \cref{sec:supp_protocols} to one shared session without specialized tools.
\textbf{ERW-Utils} supplies our constraint, scene-graph, and diagnostic tools in a single session without role prompts.
\textbf{ERW-NoCoord} combines both prompts and tools in one shared session, leaving execution order to the agent.
\textbf{EnRealWorld} adds isolated role contexts, stage-dependent information boundaries, and Manager-chaired deliberation.

\paragraph{Results.}
As shown in \cref{tab:supp_ablation}, \textbf{ERW-Prompts} improves clearly over Base Agent and achieves the highest geometry score, indicating that explicit specification, self-testing, and independent review perspectives help even within a single session.
\textbf{ERW-Utils} brings minimal gain, as an unguided session lacks clear tool-use criteria and introduces invocation noise.

Combining prompts and tools in one shared session, \textbf{ERW-NoCoord} slightly trails \textbf{ERW-Prompts}.
Mixing multi-role instructions and tool outputs in a single context causes interference and lets the session shortcut verification without stage boundaries.
With isolated role contexts and structured deliberation, \textbf{EnRealWorld} avoids this interference, achieving the highest overall score ($53.55$) and a large gain in functional completion ($53.2$ vs.\ $39.7$).

\begin{table}[t]
\centering
\caption{
\textbf{PhysCodeGym results under GPT-6.1 Sol (43 counterfactual tasks, 0--1 scale).}
Means cover all 43 tasks, counting candidate failures as zero.
Best results in each column are \textbf{bolded}.
}
\label{tab:supp_physcodegym}
\footnotesize
\setlength{\tabcolsep}{3pt}

\definecolor{v2wOurs}{RGB}{228,239,247}

\begin{tabular*}{\columnwidth}{@{\extracolsep{\fill}}lrrrr@{}}
\toprule
Method & $N$ & Visual $\uparrow$ & Physics $\uparrow$ & Failures $\downarrow$ \\
\midrule
Base Agent & $43$ & $0.336$ & $0.321$ & $19$ \\
\rowcolor{v2wOurs}\textbf{EnRealWorld} & $43$ & $\mathbf{0.491}$ & $\mathbf{0.416}$ & $\mathbf{6}$ \\
\bottomrule
\end{tabular*}
\end{table}

\begin{table}[t]
\centering
\caption{
\textbf{Video2World component ablations under GPT-6.1 Sol (39 tasks, 100-point scale).}
Endpoints are from \cref{tab:video2world_benchmark}, and $\Delta S$ reports gains over Base Agent.
\textbf{Bold} and \underline{underline} denote the best and second-best values in each column.
}
\label{tab:supp_ablation}
\footnotesize
\setlength{\tabcolsep}{3pt}

\definecolor{v2wBase}{RGB}{232,232,232}
\definecolor{v2wOurs}{RGB}{228,239,247}

\begin{tabular*}{\columnwidth}{@{\extracolsep{\fill}}lrrrrr@{}}
\toprule
Method & $S\uparrow$ & $\Delta S$ & $F\uparrow$ & $G\uparrow$ & $D\uparrow$ \\
\midrule
\rowcolor{v2wBase}Base Agent & $46.00$ & -- & $26.8$ & $64.5$ & $46.7$ \\
ERW-Prompts & \underline{$51.46$} & \underline{$+5.46$} & \underline{$39.7$} & $\mathbf{67.6}$ & $47.0$ \\
ERW-Utils & $46.75$ & $+0.75$ & $31.8$ & $62.8$ & $45.6$ \\
ERW-NoCoord & $50.67$ & $+4.67$ & $38.8$ & \underline{$65.8$} & $\mathbf{47.4}$ \\
\midrule
\rowcolor{v2wOurs}\textbf{EnRealWorld} & $\mathbf{53.55}$ & {\boldmath$+7.55$} & $\mathbf{53.2}$ & $60.1$ & \underline{$47.3$} \\
\bottomrule
\end{tabular*}
\end{table}

\subsection{Open-ended World Creation}

We use one-sentence prompts for laboratory, factory, and home scenes, with other details unspecified.
Painting inputs comprise \emph{Interior with the Artist's Easel} and \emph{House by the Railroad}.
Architectural inputs comprise 17 plans and elevations for three buildings from the Historic American Buildings Survey collection~\cite{locHABS}.

\paragraph{Scene scale and functional layout.}
EnRealWorld produces larger spatial extents and more instances in all three scenes (\cref{tab:open_scene_scale}).
Instance counts alone obscure functional differences: 246 of Base Agent's 368 home-scene instances are rug tassels and floor strips.
EnRealWorld builds a two-story residence connected by stairs, along with multi-zone layouts for the laboratory and factory.
These cases show that EnRealWorld infers functional zones from brief prompts and organizes how they connect.

\begin{table*}[tb]
\centering
\caption{
\textbf{Scene scale and functional layout from one-sentence prompts.}
Dimensions are axis-aligned bounding-box extents ($X\times Y\times Z$) excluding background geometry.
Instances are counted from Blender's evaluated scene.
}
\footnotesize
\label{tab:open_scene_scale}
\begin{tabular*}{\textwidth}{llrcp{9.2cm}}
\toprule
Scene & Method & Instances & Dimensions (m) & Functional layout \\
\midrule
Laboratory & Base Agent & 234 & $10.4\times8.1\times3.4$ & Single-story laboratory with a central bench, a sink, and wall-side storage cabinets. \\
 & EnRealWorld & 1,430 & $28.2\times20.2\times3.4$ & Single-story, multi-zone laboratory with a shared corridor connecting sample preparation, testing, heat treatment, and storage areas. \\
\midrule
Factory & Base Agent & 279 & $11.6\times8.2\times4.0$ & Single-story packaging cell with a conveyor, a robotic arm, and a control cabinet. \\
 & EnRealWorld & 3,841 & $36.9\times74.8\times13.2$ & Single-story industrial warehouse with four assembly stations, a material preparation area, a quality-inspection bench, and storage racks. \\
\midrule
Home & Base Agent & 368 & $9.5\times7.0\times3.2$ & Single-story living room with a sofa, a coffee table, chairs, and storage cabinets. \\
 & EnRealWorld & 400 & $17.8\times23.6\times7.9$ & Two-story residence with living, dining, and kitchen areas on the lower floor and stairs connecting to the upper floor. \\
\bottomrule
\end{tabular*}
\end{table*}

\begin{figure*}[t]
    \centering
    \includegraphics[width=\linewidth]{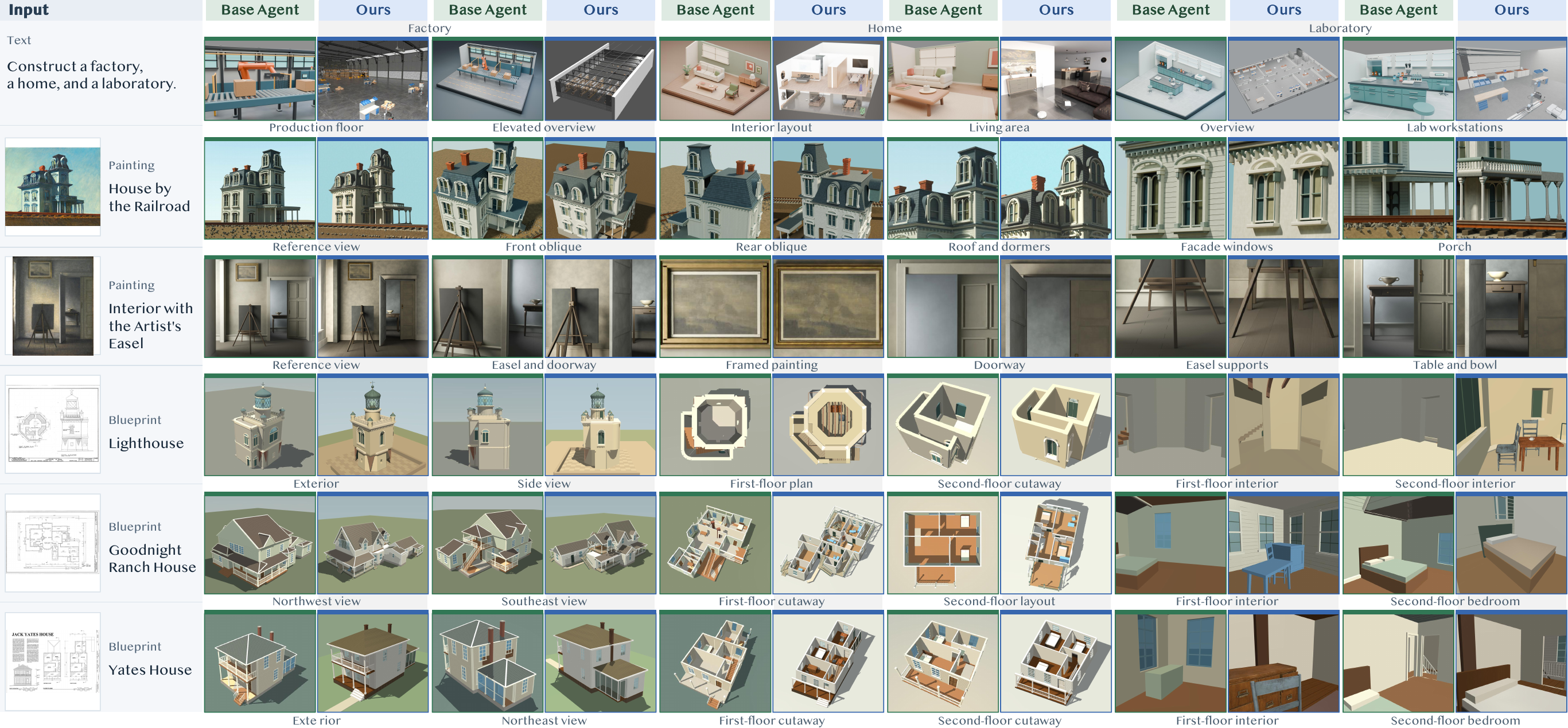}
    \caption{
    \textbf{Open-ended world creation from text, paintings, and architectural drawings.}
    Pairs compare Base Agent (left, green) with EnRealWorld (right, blue).
    The top row shows factory, home, and laboratory scenes from one-sentence prompts.
    Other rows show painting and architectural reconstructions through reference views, alternative viewpoints, details, and floor cutaways.
    }
    \label{fig:openended_world}
\end{figure*}

\paragraph{Reconstruction and spatial completion.}
For \emph{House by the Railroad}, EnRealWorld preserves the tracks, tower, and chimneys while extending dormer details to the rear facade, which Base Agent leaves largely plain (\cref{fig:openended_world}).
Architectural examples also show completion across floors.
Side and rear views reveal fewer floating or interpenetrating components in EnRealWorld's outputs.
EnRealWorld better preserves spatial relationships and plausible support in new views while maintaining the reference-view composition. Additional multi-view and orthographic comparisons appear in \cref{sec:supp_qualitative}.

\begin{figure}[t]
    \centering
    \includegraphics[width=\linewidth]{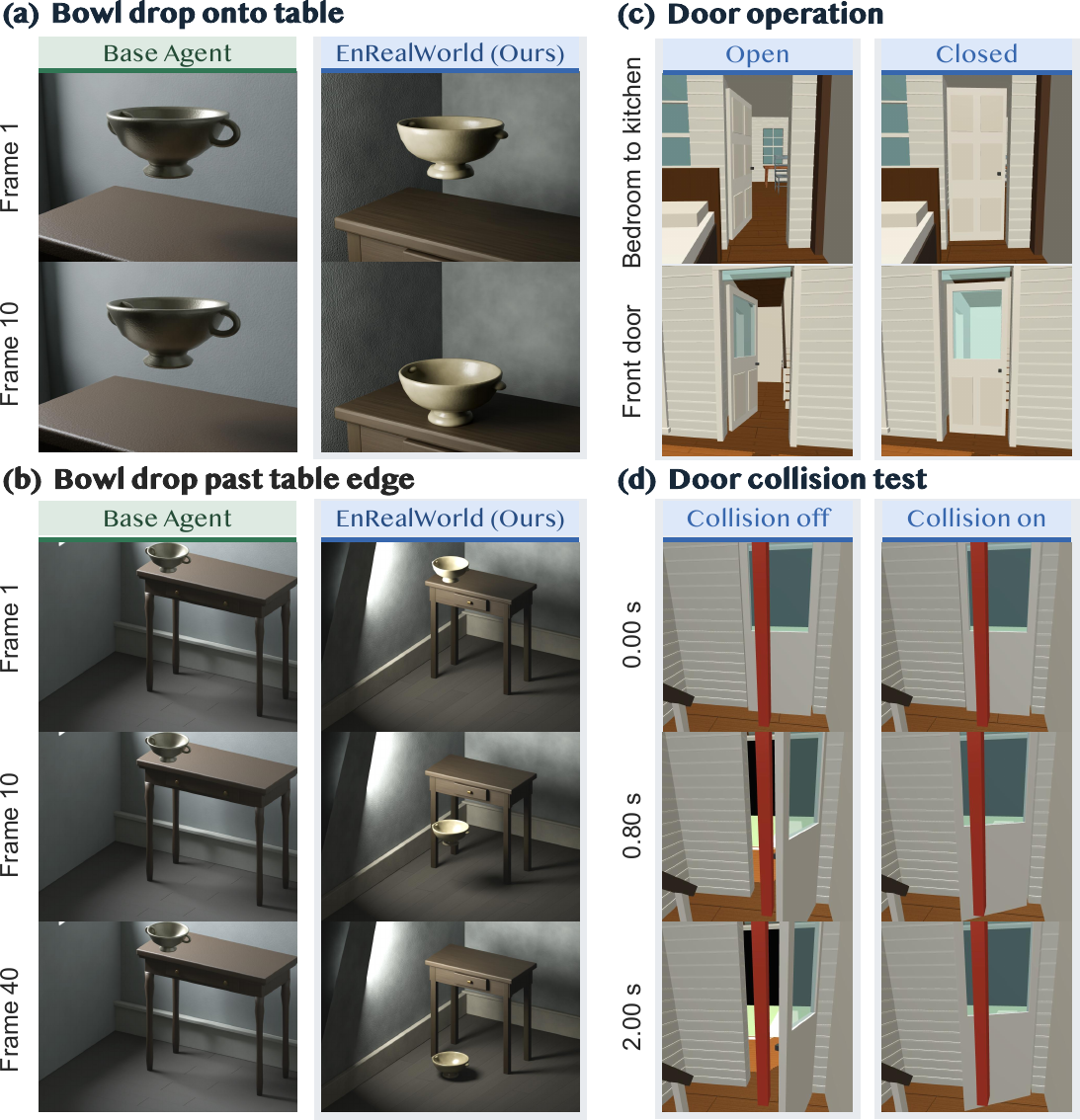}
    \caption{
    \textbf{Physical responses and interactions.}
    (a,b) A dropped bowl remains suspended with Base Agent but lands with EnRealWorld.
    (c) EnRealWorld's two doors open and close on command.
    (d) The obstacle blocks the door only with collision enabled.
    }
    \label{fig:physical_responses}
\end{figure}

\paragraph{Physical responses and sequential interaction.}
In the reconstructed studio, as illustrated in \cref{fig:physical_responses}, we release a bowl above the table and beyond its edge. 
After multiple frames, Base Agent's bowl remains suspended in both conditions, whereas EnRealWorld's bowl lands on the table or floor depending on support.
In \emph{Jack Yates House}, Base Agent lacks controllable door joints.
Doors constructed by EnRealWorld execute sequential commands across components and respond to obstacles when collision is enabled.
We further test roof support collision, variable gravity, surface-dependent friction, multi-body rack collapse, and deformable landing in \cref{sec:supp_g1}.

\subsection{Experience Accumulation and Transfer}
\label{sec:experience_transfer}
We run ten rounds of a square-table leg assembly task from Video2World.
Each round starts with an empty project and fresh agent sessions, retaining only consolidated role-specific experience with model parameters fixed.
After round ten, we freeze the library and test transfer to another square-table task and a round-table task, covering new instances and structural changes within the same task family.

\begin{figure}[t!]
    \centering
    \includegraphics[width=\linewidth]{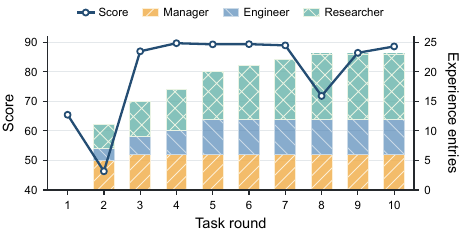}
    \caption{
    \textbf{Experience accumulation over ten rounds.}
    The line shows execution scores, and the stacked bars show role-specific library sizes at the start of each round.
    Libraries support additions, removals, and revisions.
    }
    \label{fig:experience_accumulation}
    
\par\medskip
    \includegraphics[width=\linewidth]{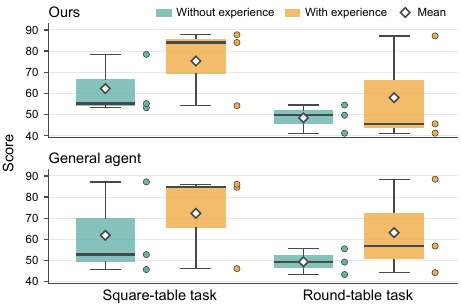}
    \caption{
    \textbf{Cross-task experience transfer across agents.} 
    Performance is evaluated on square-table (left) and round-table (right) assembly using EnRealWorld (top) versus a general-purpose coding agent (bottom), with 3 runs per condition.
    The experience library is frozen after round 10. 
    Box plots show execution scores, circles individual runs, and white diamonds indicate means.
    }
    \label{fig:experience_transfer}
\end{figure}

\paragraph{Experience accumulation.}
Library growth slows after an initial expansion (\cref{fig:experience_accumulation}).
Entry counts level off first for Manager memory, then for Engineer tools, while the Researcher continues adding runtime cases.
Execution scores increase from 65.45 in round one to 88.55 in round ten, averaging 80.26.
Performance remains non-monotonic.
Round eight scores 71.81 and produces a tilted table leg, showing that retained experience does not eliminate execution failures.
Effective reuse therefore requires checking when a retained procedure applies and validating its instantiation on the current task.

\paragraph{Cross-task transfer.}
We evaluate EnRealWorld and a general-purpose coding agent with and without the frozen library, using three independent runs per task and condition.
Experience improves mean execution scores for both agents on both tasks (\cref{fig:experience_transfer}).
The gains differ across tasks. 
Similar object structures and assembly procedures offer relatively direct opportunities to reuse construction, diagnosis, and verification methods.

\subsection{Effectiveness with Frontier Models}
\label{sec:frontier_models}

\begin{figure*}[t]
    \centering
    \includegraphics[width=\linewidth]{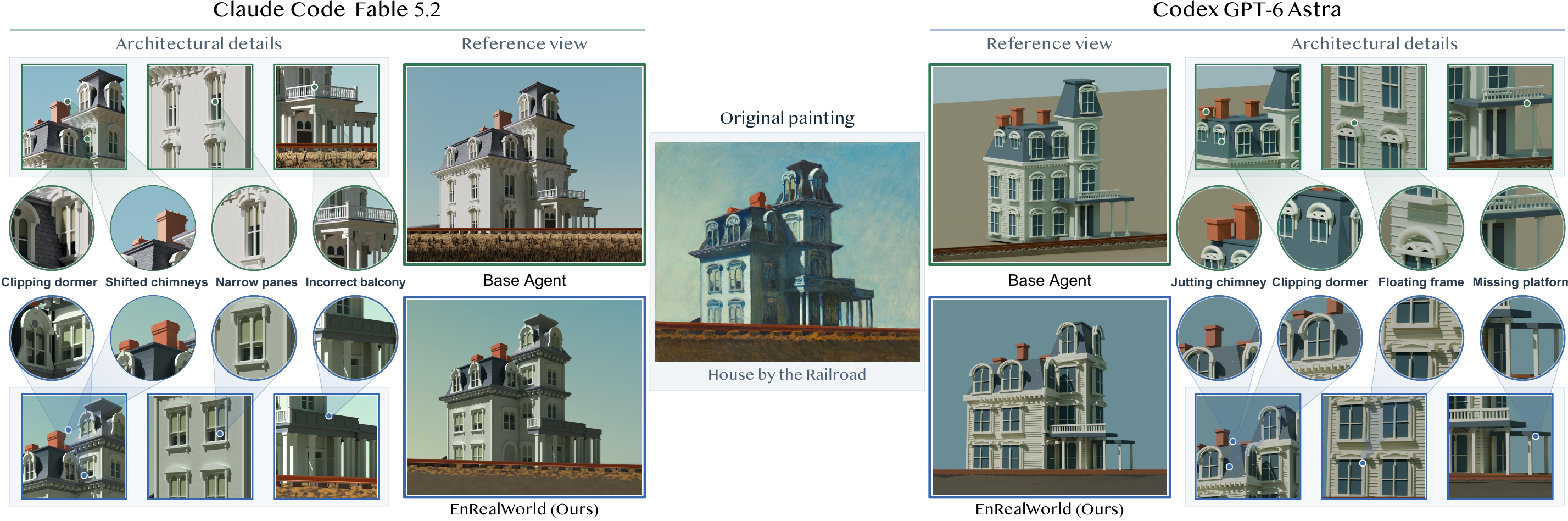}
    \caption{
    \textbf{Frontier-model results on \emph{House by the Railroad} (Claude Fable 5.2 and GPT-6 Astra).}
    Claude Code with Claude Fable 5.2 (left) and Codex with GPT-6 Astra (right) appear alongside the reference painting in the center.
    Each group compares Base Agent (top, green) with EnRealWorld (bottom, blue), and close-ups highlight localized architectural and material details.
    }
    \label{fig:frontier_models}
\end{figure*}

\begin{figure*}[tp]
    \centering
    \includegraphics[width=\textwidth,height=.80\textheight,keepaspectratio]{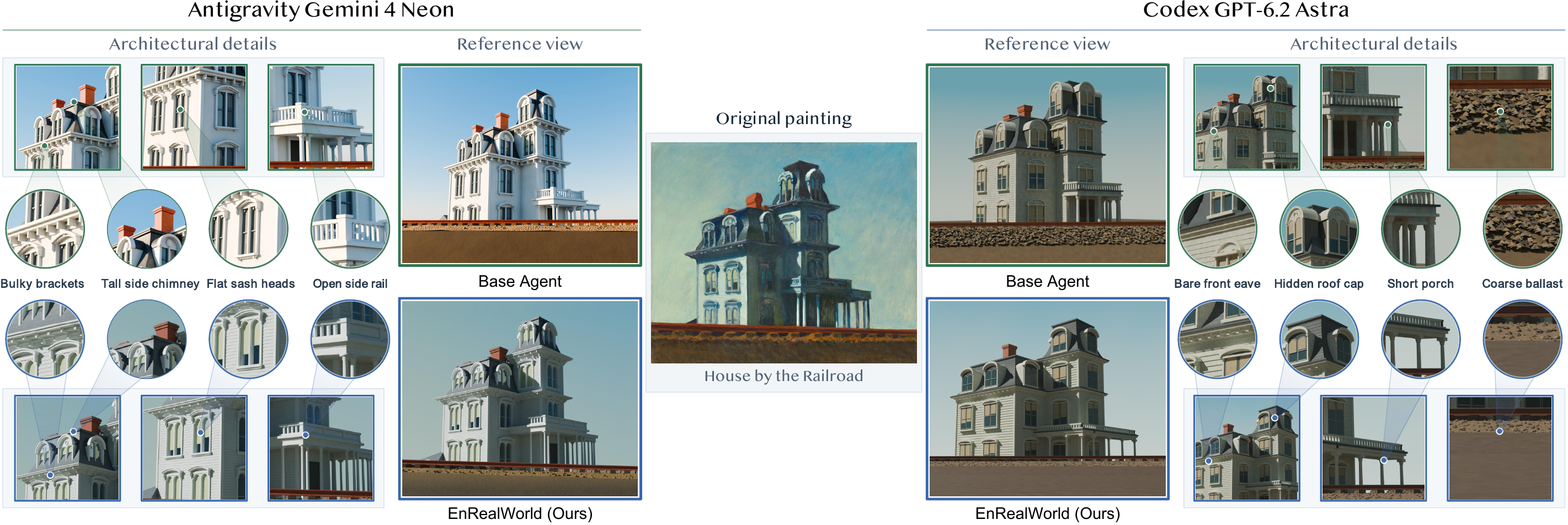}
    \caption{
    \textbf{Frontier-model results on \emph{House by the Railroad} (Gemini 4 Neon and GPT-6.2 Astra).}
    Antigravity with Gemini 4 Neon (left) and Codex with GPT-6.2 Astra (right) appear alongside the reference painting in the center.
    Each group compares Base Agent (top, green) with EnRealWorld (bottom, blue), and close-ups highlight localized architectural and material details.
    }
    \label{fig:supp_g2}
\end{figure*}

We evaluate EnRealWorld across four frontier setups on \emph{House by the Railroad} as shown in \cref{fig:frontier_models,fig:supp_g2}.

\paragraph{Reference fidelity.}
Across these setups, Base Agent generates plausible local details but drifts from the painting's lighting, proportions, and silhouette.
It alters chimney placement and foreground contrast under Claude Fable 5.2 and Gemini 4 Neon, omits or truncates porch structures under the Astra models, and renders coarse track ballast.
EnRealWorld preserves the painting's tonal balance, porch geometry, and graded rail embankment.

\paragraph{Local structural plausibility.}
Close-ups reveal geometric defects in Base Agent's outputs that materials and occlusion hide in wide views, such as window--roof interpenetration, floating frames, and missing eave or roof-cap trim.
EnRealWorld produces cleaner component boundaries and attachments across models.
Local mesh checks under Claude Fable 5.2 confirm no intersection between the window structure and the main roof.

\paragraph{Physical executability.}
In physical tests, both methods pass basic support and collision checks under Fable 5.2, while EnRealWorld pairs those behaviors with higher geometric consistency.
Under GPT-6 Astra, however, probes pass through the visible roof, rear wall, and ground in Base Agent's scene, whereas EnRealWorld's corresponding structures support or block them.
Roof-collision and surface-friction measurements appear in \cref{sec:supp_g1}.

\section{Related Work}
\label{sec:related}

\paragraph{World Models and Representations.}
Video world models~\cite{bruce2024genie} and explicit 3D representations~\cite{kerbl20233dgs} advance generation and reconstruction, but visual realism does not ensure physical correctness~\cite{motamed2026physicsiq}.
PhysGaussian~\cite{xie2024physgaussian} integrates physical dynamics, while URDFormer~\cite{chen2024urdformer} recovers articulated simulation assets.
Asking the World~\cite{zeng2026askingtheworld} calibrates executable worlds from video and selects physical probes for question answering.
We focus on constructing and revising executable worlds against task requirements.

\paragraph{Agentic World Construction.}
Agentic scene construction extends from layout synthesis~\cite{yang2024holodeck,sun2025layoutvlm} to executable and dynamic worlds.
VIGA~\cite{yin2026viga} interleaves code generation, rendering, and inspection, SceneWeaver~\cite{sceneweaver} combines extensible tools with self-reflective planning, and SceneSmith~\cite{pfaff2026scenesmith} collaboratively builds simulation-ready environments.
Code2Worlds~\cite{zhang2026code2worlds} programs dynamic worlds, while Code2Worlds-2~\cite{zhang2026code2worlds2} adds simulator compilation and feedback-driven refinement.
SimWorlds~\cite{liu2026simworlds} uses stage-wise mechanism checks, and WorldAgent~\cite{wang2026worldagent} revises physical worlds using geometry and simulation checks.
Reflexion~\cite{shinn2023reflexion} supports experience reuse through linguistic reflections, and Voyager~\cite{wang2023voyager} through executable skills.
EnRealWorld couples revisable and implementation-independent specifications with independently designed physical interventions.
Stage-dependent information boundaries coordinate construction and verification, while role-specific experience guides subsequent tasks.

\section{Conclusion}

We presented EnRealWorld, an agentic engineering framework that couples specification refinement, world construction, proactive investigation, and experience accumulation.
Stage-dependent information boundaries enable complementary roles to construct and independently verify executable worlds.
Experiments show improvements in task completion, reference fidelity, and physical responses, together with experience reuse across related tasks and agent configurations.
Several directions remain open.
Budget-aware scheduling and reuse of verified components could reduce the token and latency costs of verification and deliberation.
Continual interactive editing could incorporate user modifications, new requirements, and incoming observations while preserving verified behaviors.
Experience reuse could further improve through applicability checks and adaptation into verifiable operations.


\clearpage
{
    \small
    \bibliographystyle{ieeenat_fullname}
    \bibliography{main}
}

\clearpage
\setcounter{page}{1}
\renewcommand{\thepage}{S\arabic{page}}
\maketitlesupplementary

\appendix

\setcounter{table}{0}
\renewcommand{\thetable}{S\arabic{table}}
\renewcommand{\theHtable}{supp.\arabic{table}}
\setcounter{figure}{0}
\renewcommand{\thefigure}{S\arabic{figure}}
\renewcommand{\theHfigure}{supp.\arabic{figure}}
\setcounter{equation}{0}
\renewcommand{\theequation}{S\arabic{equation}}
\renewcommand{\theHequation}{supp.\arabic{equation}}
\setcounter{algorithm}{0}
\renewcommand{\thealgorithm}{S\arabic{algorithm}}
\providecommand{\theHalgorithm}{}
\renewcommand{\theHalgorithm}{supp.\arabic{algorithm}}

\AtBeginEnvironment{promptbox}{\par\ifdefined\nolinenumbers\nolinenumbers\fi}
\newtcolorbox{promptbox}[1]{
  breakable,
  enhanced,
  lines before break=3,
  title after break={#1 (continued)},
  colback=black!3,
  colframe=black!22,
  colbacktitle=black!10,
  coltitle=black,
  boxrule=0.5pt,
  titlerule=0.4pt,
  arc=1pt,
  left=6pt,
  right=6pt,
  top=5pt,
  bottom=5pt,
  toptitle=3pt,
  bottomtitle=3pt,
  before skip=6pt,
  after skip=6pt,
  fonttitle=\footnotesize\bfseries,
  fontupper=\footnotesize,
  title={#1}
}

\section{Supplementary Formulation}
\label{sec:supp_formulation}
We expand the deliverables, two-timescale dynamics, quality objective, and experience update introduced in \cref{sec:formulation}. 

\subsection{Task Representation and Deliverables}
\label{sec:supp_a1}
A 4D world creation task is given by $x=(u,\mathcal{O},\mathcal{A})$, comprising natural-language requirements $u$, multimodal observations $\mathcal{O}$, and available assets with metadata $\mathcal{A}$.
Any component may be absent for a given task.
Observations typically cover only a visible view or a short motion segment, leaving responses to novel interventions underdetermined.
An executable world $W=(\mathcal{G},\mathcal{P},\mathcal{I})$ encodes entities, geometry, and appearance ($\mathcal{G}$), physical models, constraints, and parameters ($\mathcal{P}$), and control logic and interaction interfaces ($\mathcal{I}$).
For example, a single painting constrains the static placement of a bowl on a tabletop, while releasing the bowl past the table edge tests support and collision mechanics outside the depicted view.

Task $n$ returns the deliverable tuple $\mathcal{Y}_n=(W_n^{\mathrm{out}},\mathcal{B}_n,\mathcal{R}_n)$, where $W_n^{\mathrm{out}}$ packages the runnable world and its dependencies, $\mathcal{B}_n$ stores process artifacts such as candidate snapshots, simulation traces, and probe logs, and $\mathcal{R}_n$ records per-check outcomes and unresolved issues.
When construction fails to yield a runnable world, the system returns $W_n^{\mathrm{out}}=\varnothing$ together with $\mathcal{B}_n$ and a failure report in $\mathcal{R}_n$.

\subsection{Two-Timescale Construction Dynamics}
\label{sec:supp_a2}
World creation spans two nested timescales. 
At the outer construction step $k$, the state $z_k=(\mathcal{C}_k,W_k,\mathcal{F}_k)$ tracks the specification $\mathcal{C}_k$, candidate world $W_k$, and accumulated observations and test outcomes $\mathcal{F}_k$. 
Conditioned on $x$, $z_k$, and persistent experience $\mathcal{M}$, the construction policy $\pi$ selects an operation $d_k$ and the state transition kernel $\mathcal{T}$ updates the state to $z_{k+1}$:
\begin{equation}
d_k \sim \pi(\cdot \mid x, z_k, \mathcal{M}), \qquad z_{k+1} \sim \mathcal{T}(\cdot \mid z_k, d_k, x).
\end{equation}

Clarification refines $\mathcal{C}_k$, construction and repair revise $W_k$, and observations or probes expand $\mathcal{F}_k$. One construction decision may trigger a full simulation rollout indexed by $\tau \in \{0,\dots,T-1\}$ under backend $b$ and runtime configuration $\eta$:
\begin{equation}
s_{\tau+1} = f_{W,b}(s_\tau, a_\tau; \eta), \qquad o_\tau = g_{W,b}(s_\tau),
\end{equation}
where $T$ is the rollout horizon, $s_\tau$ and $a_\tau$ denote the simulator state and applied action or intervention at step $\tau$, $f_{W,b}$ is the dynamics transition under world $W$ and backend $b$, and $g_{W,b}$ maps $s_\tau$ to the rendered or sensor observation $o_\tau$. 
Under partial observability, identical rendered or sensor observations $o_\tau$ can correspond to distinct underlying contact states $s_\tau$.

Each check $c_i=(\mathcal{D}_i,\phi_i)$ specifies initial conditions and actions or interventions ($\mathcal{D}_i$) along with a predicate ($\phi_i$) over trajectories, observations, or simulator states.
Executed checks return \emph{Pass} or \emph{Fail}. 
Checks blocked by missing tools, unavailable readouts, or budget exhaustion are recorded as \emph{Unverified}.
Revisions trigger targeted retesting of affected behaviors.
Construction ends when all mandatory delivery criteria hold or the execution budget runs out, with $\mathcal{R}_n$ recording every check outcome.

\subsection{Feasible Space and Quality Objective}
\label{sec:supp_a3}
The search space $\Omega$ combines discrete structural choices, such as door articulation, with continuous parameters like hinge pose, mass, and friction.
Mandatory requirements $\mathcal{C}_{\mathrm{hard}}$ define the feasible set $\Omega_{\mathrm{valid}}(x,\mathcal{C}_{\mathrm{hard}})\subseteq\Omega$. 
Under evaluation conditions $\mathcal{D}_{\mathrm{eval}}$, an ideal world $W^\star$ minimizes a conceptual quality loss $\mathcal{L}$:
\begin{equation}
W^\star \in \arg\min_{W \in \Omega_{\mathrm{valid}}(x,\mathcal{C}_{\mathrm{hard}})} \mathcal{L}(W; x, \mathcal{D}_{\mathrm{eval}}),
\end{equation}
where $\mathcal{L}=\sum_{j\in\{f,p,r\}}\lambda_j\mathcal{L}_j(W;x,\mathcal{D}_{\mathrm{eval}})$ with $\lambda_j\geq 0$. The three terms measure input fidelity ($f$), physical consistency ($p$), and interaction robustness ($r$).
Input fidelity measures structural and dynamic agreement with input observations and requirements under reference conditions. 
Physical consistency measures compliance with rigid-body dynamics, contact laws, and kinematic constraints in the chosen simulation regime. 
Interaction robustness measures response errors under prescribed changes to initial states, action sequences, or external interventions, where valid physical responses depart from reference trajectories. 
During construction, agents act on executable checks and diagnostic feedback directly, without optimizing a scalar surrogate.

\subsection{Cross-Task Experience Evolution}
\label{sec:supp_a4}
At task $n$, the persistent experience state $\mathcal{M}_n$ conditions the construction policy $\pi$ while base model parameters $\theta$ remain fixed.
After task completion, reusable information extracted from process artifacts $\mathcal{B}_n$, including internal observations, execution outcomes, and test logs, drives the experience update operator $\mathcal{U}$ to produce $\mathcal{M}_{n+1} = \mathcal{U}(\mathcal{M}_n, \mathcal{B}_n)$, expanding \cref{sec:formulation-experience}.
Updates retain, extend, revise, or discard entries from internal records alone.
External benchmark scores are used only for reporting and never enter $\mathcal{U}$.
In cross-task studies such as \cref{sec:experience_transfer}, every task starts from an empty workspace so that only transferable procedures, reusable tools, and diagnostic records persist across tasks.
In the ten-round study, transfer uses the library after the tenth update with no further updates.

\section{Supplementary Methodology}
\label{sec:supp_method}

This section provides concrete SMT queries and runtime role prompts complementing \cref{sec:method}.

\subsection{Specification Checking with SMT}
\label{sec:supp_b2}
Complementing \cref{sec:method-proposition}, we illustrate Z3 reasoning with a specification-conflict check by the Manager and an archived candidate review by the Researcher.

\paragraph{Illustrative requirement-conflict resolution (Manager).}
Consider a passage-width specification in meters where an explicit requirement enforces a minimum width of $0.90\,\mathrm{m}$ while an initial image estimate suggests an upper bound of $0.80\,\mathrm{m}$.
Incremental solving (\texttt{push}/\texttt{pop}) detects the contradiction under the inferred bound and confirms consistency once the bound is retracted:
\begin{lstlisting}[breaklines=true,breakatwhitespace=false,columns=fullflexible,keepspaces=true,basicstyle=\ttfamily\scriptsize]
(declare-const width Real)
(assert (>= width 0.90))
(push 1)
(assert (<= width 0.80))
(check-sat) ; unsat
(pop 1)
(check-sat) ; sat
\end{lstlisting}
The initial \texttt{unsat} result prompts scale remeasurement. 
Afterward, the Manager records the measured bound with its source while keeping the mandatory lower bound. 
Physical clearance and collision still require simulation tests.

\paragraph{Archived candidate-review disambiguation (Researcher).}
During candidate review, the Researcher can use Z3 to test auxiliary interpretations against observed and planned trajectories. 
In the square-table leg assembly task, which permits an independently planned robot trajectory, the first independent review checked whether the candidate's initial action depth \texttt{action0\_z}, bounded within $\pm 10\,\mathrm{mm}$ of the declared part depth, could match the video-estimated tool-center depth \texttt{video0\_z}, which carries $\pm 30\,\mathrm{mm}$ uncertainty. Exported in meters at double precision, the archived SMT-LIB2 query reads:
\begin{lstlisting}[breaklines=true,basicstyle=\ttfamily\scriptsize]
(declare-const action0_z Real)
(declare-const video0_z Real)
(declare-const same_initial_required Bool)
(assert (and (<= 0.8384818920211649 action0_z) (<= action0_z 0.8584818920211649)))
(assert (and (<= 0.7415455293655396 video0_z) (<= video0_z 0.8015455293655396)))
(assert (=> same_initial_required
            (and (<= (- video0_z 0.01) action0_z) (<= action0_z (+ video0_z 0.01)))))
(assert (not same_initial_required))
(check-sat)
\end{lstlisting}
Evaluating this query in Z3 returns \texttt{sat}. Replacing the penultimate line with \texttt{(assert same\_initial\_required)} returns \texttt{unsat}, since the two intervals are separated by at least $36.94\,\mathrm{mm}$ against the $10\,\mathrm{mm}$ tolerance. Guided by this result, the Researcher retained the authorized self-planned path and isolated a separate motion discrepancy for Engineer repair: at frame 28, the candidate leg center projected approximately $19$ pixels left of the source landmark. The first review remained marked as \emph{Fail}, and the projection discrepancy was assigned to the Engineer for repair.

\subsection{Agent Role and Workflow Prompts}
\label{sec:supp_protocols}
We provide a reusable EnRealWorld skill that users can ask their coding agent to follow to prepare task inputs and launch the three-role workflow.
For readability, the seven boxes condense the instructions used at runtime. The first four summarize the shared protocol and each role's responsibilities. The last three show what each role is asked to do at each stage. On the first turn, the runner supplies the task, shared protocol, phase instructions, and paths to the role skill and shared constraint guidance. On resumed turns, it repeats the shared protocol and current phase instructions and points to the full context, coordination records, and candidate binding when available. The phase names below are those used by the scheduler.

\begin{promptbox}{Shared Team Protocol (All Roles)}
Preserve original input values and cite sources. Distinguish native readbacks from declarations and estimates.
\par\smallskip
Try supplied data and explicit template values first. Before any substitution, retain the baseline check, concrete failure or incompatibility record, and replacement rationale.
\par\smallskip
Treat missing tools or unrun checks as \emph{Unverified}, never as \emph{Pass} or \emph{Fail}. A failed attempt does not authorize relaxing fixed constraints.
\par\smallskip
Return a JSON object with \texttt{status} (\texttt{passed} or \texttt{failed}) and a summary. Use \texttt{passed} only when evidence verifies the phase requirements. If a required check remains unverified, return \texttt{failed} for the phase and identify the untested claim as unknown. Keep the frozen candidate read-only and bind evidence to the version tested.
\end{promptbox}

\begin{promptbox}{Manager Role Prompt}
Read original inputs and maintain the proposition checklist (\emph{Must}, \emph{Can}, \emph{May}), checking for omissions, weakened constraints, and conflicts (using Z3 for formalizable conditions).
\par\smallskip
Audit the candidate, Engineer repairs, and Researcher findings against the accepted checklist and list unchecked items. If the Engineer omits the current checklist version, mark self-test coverage as unconfirmed.
\par\smallskip
Chair deliberation by querying the Engineer or Researcher as needed, continuing while a feasible distinguishing test remains within budget, and closing material findings only with verified revisions and independent retests.
\end{promptbox}

\begin{promptbox}{Engineer Role Prompt}
Build a runnable world from original inputs using test-driven development, deriving expected behavior and self-tests first.
\par\smallskip
Supply execution entry points, runtime descriptors, coordinate and unit transforms, and entity-to-native-object mappings for independent tests.
\par\smallskip
For image or video inputs, compare the target shape, scale, initial pose, and key contacts with the input before major physical tests. Use an overlay when feasible and correct discrepancies that would affect the test conclusions. Self-test against the accepted checklist after generation and repair feasible gaps.
\par\smallskip
Optionally delegate separable subtasks (assets, measurements, components, or local checks) while owning end-to-end integration and verification.
\end{promptbox}

\begin{promptbox}{Researcher Role Prompt}
Act as an independent, skeptical reviewer deriving a scene graph and test plan directly from original inputs and the seven probe families.
\par\smallskip
Evaluate frozen candidates (never mutating them) across appearance, motion, and controlled physical interventions, separating input conformance from observed visual and physical effects. If a check is reported as infeasible, inspect the supplied materials for another way to run it.
\par\smallskip
State competing explanations for material discrepancies, run distinguishing tests or Z3 counterexample checks, report requirement conflicts to the Manager, and retest affected behaviors after revisions.
\par\smallskip
Optionally delegate visual, trajectory, or physical checks under the same read-only boundary while owning conflict reconciliation and final review.
\end{promptbox}

\begin{promptbox}{Stage 1: Co-Design Phase Prompts}
\textbf{Manager (\texttt{intake} / \texttt{work}).} At intake, register sources and propositions, check for omitted or weakened requirements, and submit the Markdown checklist with the checklist tool. In later work turns, check the runner's receipt and resubmit if rejected.
\par\smallskip
\textbf{Engineer (\texttt{work}, conditional \texttt{self\_test}).} Start building and testing from the original inputs while the Manager prepares the checklist. After generation, the runner requests a checklist self-test on first acceptance or a version change. Report checked and unchecked items, evidence, and the version used in \texttt{requirement\_version}.
\par\smallskip
\textbf{Researcher (\texttt{work}).} Derive tests from the original task, add applicable tests from the seven probe families, and merge duplicates into a plan with observations and pass criteria.
\end{promptbox}

\begin{promptbox}{Stage 2: Cross-check Phase Prompts}
\textbf{Researcher (\texttt{review}).} Test the frozen candidate against the original inputs. Record input conformance separately from observed visual and physical effects, with evidence and \texttt{candidate\_digest}. Send requirement conflicts to the Manager. For a material discrepancy, state competing explanations, run a test that distinguishes them when feasible, and send the evidence and proposed repair to the Manager. Keep unresolved material findings failed and record attempted tests and blockers.
\par\smallskip
\textbf{Engineer (\texttt{repair}).} After a failed review, make one repair attempt on the working copy. Run a distinguishing test, rerun affected checks, and report evidence, checklist coverage, and any remaining blockers.
\par\smallskip
\textbf{Researcher (\texttt{retest}).} If the repair changes the delivery digest, the runner freezes the revision. Rerun the original failing checks and affected probes against the baseline, then return evidence bound to the new \texttt{candidate\_digest}.
\end{promptbox}

\begin{promptbox}{Stage 3: Deliberate Phase Prompts}
\textbf{Manager (\texttt{manager\_review}, then \texttt{meeting}).} Audit checklist coverage, input substitutions, and evidence versions. During the meeting, return a \texttt{decision} of \texttt{engineer} or \texttt{researcher} to ask that role a question. The respondent returns evidence and open issues to the Manager. Use \texttt{continue} to start another round, optionally assigning work by role, or \texttt{finish} to report the outcome. A passing finish requires the reviewed candidate digest and the current accepted checklist version previously handed to the Engineer for self-testing. Record blocked and unverified checks.
\end{promptbox}

\section{Experimental Setup}
\label{sec:supp_setup}
\subsection{Tasks and Inputs}
\label{sec:supp_c1}
Our primary benchmark evaluation uses Video2World~\cite{tang2026video2world}, covering seven embodied task families: DexYCB, DROID, HOI4D, HOT3D, In-house, OakInk, and RoboDojo. For physical dynamics, the fixed PhysCodeGym~\cite{shao2026physcodegym} release contains 156 tasks (95 rigid-body, 40 liquid, 12 cloth, and 9 wind), of which 43 specify counterfactual checks.

Open-ended evaluation covers a one-sentence prompt (\emph{``Construct a factory, a home, and a laboratory''}), two paintings (\emph{Interior with the Artist's Easel} and \emph{House by the Railroad}), and 17 architectural plans and elevations of three buildings from the Historic American Buildings Survey~\cite{locHABS}. Experience accumulation uses a square-table leg assembly task from Video2World, while cross-task transfer tests a new square-table instance and a structurally distinct round-table task.
When an input appears in multiple experiments, each evaluation runs independently under its own configuration.

\subsection{Models and Baseline Adaptation}
\label{sec:supp_c2}
Video2World experiments run in Codex with GPT-6.1 Sol, DeepSeek-V4.1-Flash-1017, and Kimi-3.1 Pro.
Within each model setting, Base Agent, VIGA~\cite{yin2026viga}, SceneSmith~\cite{pfaff2026scenesmith}, SceneWeaver~\cite{sceneweaver}, Code2Worlds-2~\cite{zhang2026code2worlds2}, and EnRealWorld share the same agent framework and model.
Main open-ended experiments use Codex with GPT-6.1 Sol, and frontier-model configurations appear in \cref{sec:frontier_models}.
Across standard EnRealWorld runs, all roles and the orchestrating session share the same agent framework and model at medium reasoning effort, starting from empty experience libraries unless specified otherwise.

Within each comparison, the agent framework and model are fixed across construction methods.
Adapting each baseline to the shared agent framework preserves its core planning, synthesis, and verification stages while standardizing task inputs and runtime environments.
Specifically, \textbf{Base Agent} executes the original task directly in a single session without multi-stage scaffolding.
\textbf{VIGA}~\cite{yin2026viga} interleaves scene code generation, rendering, and visual inspection, keeping its inverse-graphics refinement loop under the shared agent framework.
\textbf{SceneSmith}~\cite{pfaff2026scenesmith} collaboratively builds simulation-ready environments through staged layout planning, asset placement, and physics setup. \textbf{SceneWeaver}~\cite{sceneweaver} pairs extensible scene tools with self-reflective planning, retaining its tool-use and iterative revision cycle in the unified environment.
\textbf{Code2Worlds-2}~\cite{zhang2026code2worlds2} synthesizes dynamic world code with simulator compilation and feedback-driven refinement, preserving its compile-and-repair workflow under matched inputs and runtime settings.

Asset-level 3D generative pipelines such as EmbodiedGen V2\footnote{\url{https://horizonrobotics.github.io/EmbodiedGen/}} (with SAM3D, TRELLIS, or Hunyuan3D Pro backends) operate at a different deliverable granularity and introduce external generative dependencies.
We omit them here because fair comparison requires controlling for both the generative backend and the coding agent. Across open-ended tasks, agents retrieve assets autonomously from NVIDIA SimReady Assets~\cite{nvidiaSimReadyAssets} and ChemLab-Assets~\cite{song2026chemlabassets} without scene-specific preselection by the authors.
Task-dependent simulation backends include Blender 4.0.2 or 5.2.2 LTS, MuJoCo 3.11.0, SAPIEN 3.0.3 with ManiSkill 3.0.1, and Isaac Sim 5.1.

\subsection{Evaluation Protocols}
\label{sec:supp_c3}
Final evaluation is separate from construction-time self-tests.
Video2World tasks use the official evaluation pipeline, reporting overall score $S$ alongside function ($F$), geometry ($G$), and dynamics ($D$).
The overall and per-family means use the official Video2World scores.

PhysCodeGym evaluates frozen source submissions inside an isolated, network-disabled official verifier container.
The verifier rebuilds and executes the frozen scene from scratch and reports visual and physical scores separately.
The verifier assesses visual and motion fidelity by comparing rendered frames with the reference video using perceptual and pixel-level metrics (including SSIM, LPIPS, and PSNR) together with spatiotemporal motion consistency. 
Physical verification runs static and runtime validity checks, event consistency checks, and counterfactual physical probes.

Open-ended worlds are evaluated with an agent-as-judge workflow paired with human inspection.
Across competing methods on a given task, the judge uses the same model, agent framework, and reasoning effort, inspecting the delivered world directly to check visual alignment across reference and alternative viewpoints and running additional physical interventions to record responses.
Human inspection focuses on visual reconstruction details, especially inspecting the delivered project files directly for geometric fit, interior layout, and component interpenetration.
Visual inspection records also verify camera visibility.

\section{Supplementary Benchmark Results}
\label{sec:supp_bench}

\subsection{Video2World}
\label{sec:supp_d1}

\begin{figure*}[tp]
    \centering
    \includegraphics[width=\textwidth,height=.90\textheight,keepaspectratio]{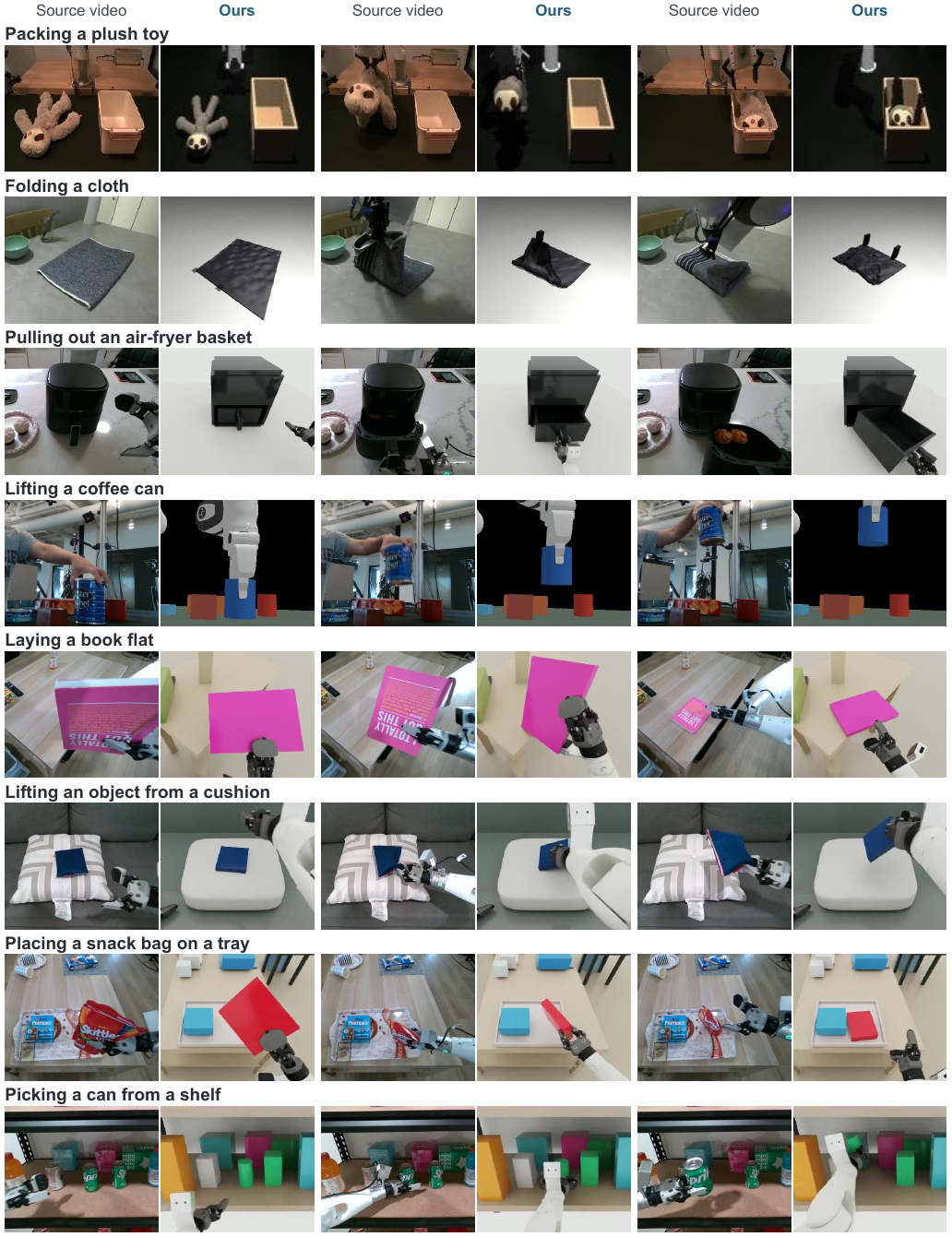}
    \caption{
    \textbf{Representative Video2World results with GPT-6.1 Sol.}
    Each row shows three temporally aligned pairs comparing the source video (left) with the executable world constructed by EnRealWorld (right, ``Ours'') across key manipulation stages.
    Following the official Video2World protocol, supported scenes use simplified object geometry to emphasize primary object motion and stage-wise manipulation progression across rigid, deformable, and articulated tasks.
    }
    \label{fig:supp_v2w_qual}
\end{figure*}

This section complements \cref{tab:video2world_benchmark} with signed per-family gains of EnRealWorld over Base Agent across all three models (\cref{tab:supp_v2w}) and representative trajectory comparisons (\cref{fig:supp_v2w_qual}).
Across all three models, overall improvements stem primarily from functional completion and dynamics on manipulation-heavy families, accompanied by modest geometry trade-offs on a subset of families.

\begin{table}[t]
\centering
\caption{
\textbf{Signed Video2World score differences (EnRealWorld minus Base Agent) across three models and seven task families.}
Each cell reports $\Delta S$ above $\Delta F \mid \Delta G \mid \Delta D$.
Positive $\Delta S$ values are bolded.
}
\label{tab:supp_v2w}
\footnotesize
\setlength{\tabcolsep}{1.2pt}

\definecolor{v2wOurs}{RGB}{228,239,247}

\newcommand{\vdelta}[2]{%
\ensuremath{\vcenter{\hbox{\shortstack[c]{#1\\[-1pt]{\scriptsize$#2$}}}}}%
}
\newcommand{\vpos}[1]{{\boldmath$#1$}}
\newcommand{\vhdr}[1]{\ensuremath{\vcenter{\hbox{#1}}}}

\begin{tabular*}{\columnwidth}{@{\extracolsep{\fill}}l|ccc@{}}
\toprule
\multicolumn{1}{c|}{\vhdr{Family}}
 & \vhdr{GPT-6.1 Sol}
 & \vhdr{\shortstack[c]{DeepSeek-V4.1-\\Flash-1017}}
 & \vhdr{Kimi-3.1 Pro} \\
\midrule
\rowcolor{v2wOurs}\textbf{Overall}
 & \vdelta{\vpos{+7.55}}{{+}26.4|{-}4.4|{+}0.6}
 & \vdelta{\vpos{+14.00}}{{+}30.2|{+}3.5|{+}8.3}
 & \vdelta{\vpos{+9.16}}{{+}25.0|{-}4.3|{+}6.9} \\
\midrule
DexYCB
 & \vdelta{\vpos{+7.63}}{{+}41.6|{-}4.8|{-}13.8}
 & \vdelta{$-17.77$}{{-}50.0|{+}4.0|{-}7.2}
 & \vdelta{$-0.97$}{{+}8.4|{-}12.2|{+}1.0} \\
\addlinespace[2pt]
DROID
 & \vdelta{$-0.24$}{{-}1.5|0.0|{+}0.8}
 & \vdelta{\vpos{+7.01}}{{+}6.6|{+}4.8|{+}9.8}
 & \vdelta{\vpos{+0.13}}{{+}11.0|{-}9.4|{-}1.2} \\
\addlinespace[2pt]
HOI4D
 & \vdelta{\vpos{+0.17}}{0.0|{-}2.3|{+}2.8}
 & \vdelta{$-3.21$}{{-}12.6|{-}4.5|{+}7.3}
 & \vdelta{\vpos{+27.45}}{{+}37.5|{+}14.0|{+}30.9} \\
\addlinespace[2pt]
HOT3D
 & \vdelta{$-9.87$}{0.0|{-}14.7|{-}15.1}
 & \vdelta{$-9.04$}{{-}18.8|{-}7.4|{-}0.9}
 & \vdelta{\vpos{+12.78}}{{+}18.8|{+}22.9|{-}3.3} \\
\addlinespace[2pt]
In-house
 & \vdelta{\vpos{+18.62}}{{+}59.2|{-}8.4|{+}5.2}
 & \vdelta{\vpos{+29.06}}{{+}70.3|{+}5.0|{+}11.9}
 & \vdelta{\vpos{+19.27}}{{+}47.8|{-}3.6|{+}13.6} \\
\addlinespace[2pt]
OakInk
 & \vdelta{$-2.28$}{{-}6.3|{+}2.5|{-}3.0}
 & \vdelta{\vpos{+1.13}}{0.0|{-}5.7|{+}9.1}
 & \vdelta{$-7.50$}{{-}12.5|{-}13.1|{+}3.2} \\
\addlinespace[2pt]
RoboDojo
 & \vdelta{\vpos{+13.62}}{{+}46.2|{-}2.8|{-}2.6}
 & \vdelta{\vpos{+34.90}}{{+}96.2|{+}11.7|{-}3.2}
 & \vdelta{\vpos{+1.47}}{0.0|{-}3.8|{+}8.3} \\
\bottomrule
\end{tabular*}
\end{table}

\section{Supplementary Open-Ended World Creation Results}
\label{sec:supp_qualitative}

\subsection{Worlds from a One-Sentence Prompt}
\label{sec:supp_f1}

\begin{figure*}[tp]
    \centering
    \includegraphics[width=\textwidth,height=.89\textheight,keepaspectratio]{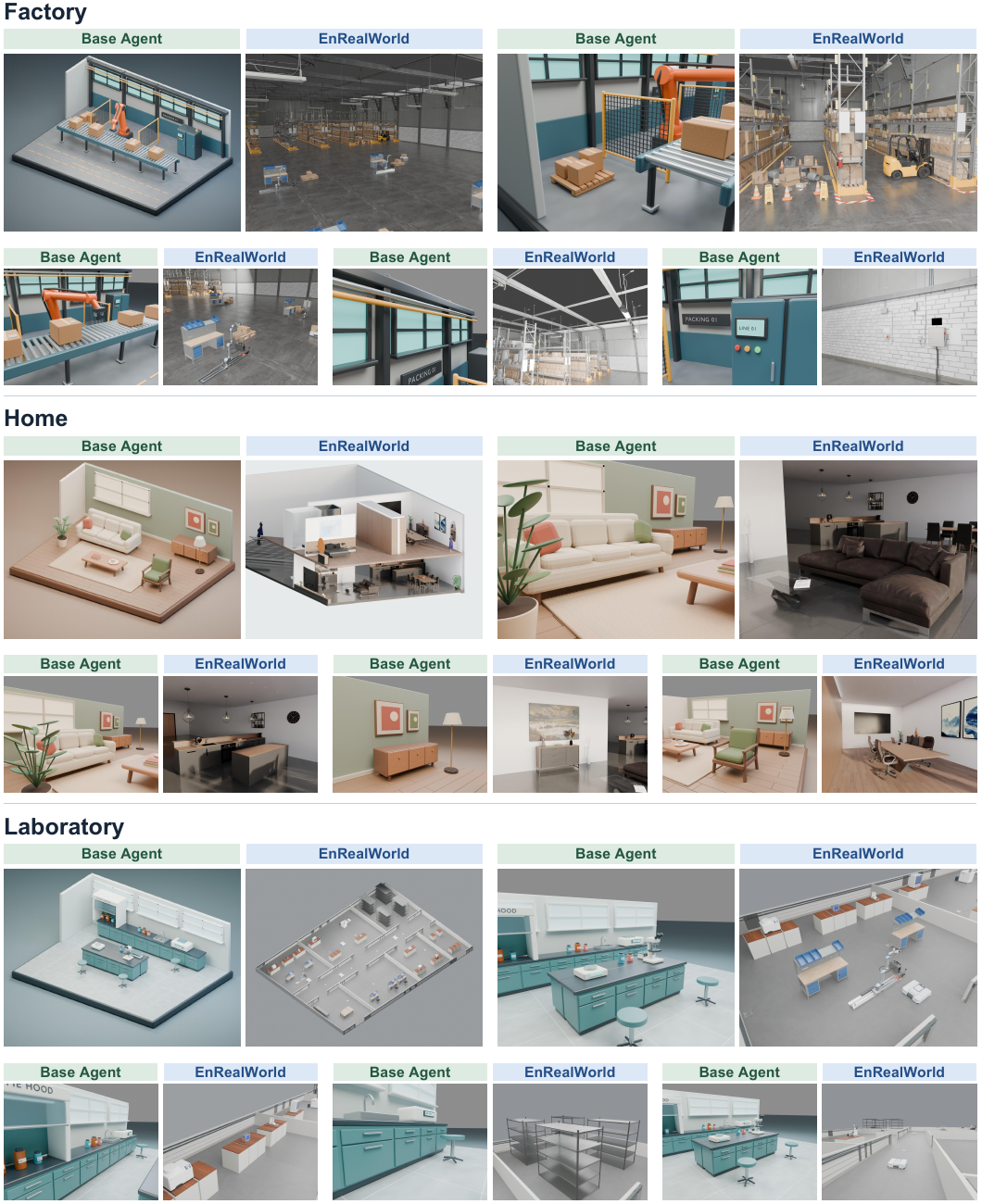}
    \caption{
    \textbf{Detailed comparisons for worlds generated from a one-sentence prompt.} 
    For \emph{Factory}, \emph{Home}, and \emph{Laboratory}, each group pairs Base Agent (left, green) with EnRealWorld (right, blue) across two wide views (top row) and three sub-zone or structural detail views (bottom row).
    Whereas Base Agent confines each scene to a single room, EnRealWorld infers full architectural scale, multi-zone or multi-story layouts, inter-zone circulation, and domain-specific equipment and infrastructure.
    }
    \label{fig:supp_f1}
\end{figure*}

\cref{fig:supp_f1} provides expanded qualitative comparisons for the three scenes generated from the one-sentence prompt, complementing \cref{fig:openended_world}.
For each scene, we show five paired views comparing Base Agent (left) and EnRealWorld (right), with wide architectural and primary-zone views in the top row and closer sub-zone and fixture views in the bottom row.
Across all three scenes, Base Agent produces a compact single-room vignette repeated across viewpoints, while EnRealWorld builds multi-zone or multi-story layouts with realistic architectural scale and infrastructure.

\paragraph{Factory.}
Base Agent constructs a small single-room packaging cell consisting of one conveyor belt, a robotic arm, a safety fence, a single stacked pallet, and a wall-mounted control panel.
EnRealWorld builds a full-scale enclosed industrial warehouse featuring steel roof trusses, overhead high-bay lighting, wide concrete aisles, multi-tier pallet racking with a forklift and safety cones, modular workstations with rail-mounted manipulators, and wall-mounted electrical infrastructure.

\paragraph{Home.}
Base Agent generates a single living room containing a sofa, coffee table, armchair, and console cabinet.
Although its instance count appears comparable to EnRealWorld's (368 vs.\ 400), 246 of those instances are repetitive rug tassels and floor strips, and all five viewpoints remain confined to the same room.
EnRealWorld creates a two-story residence with an open-plan lower floor integrating living, kitchen, and dining zones, connected via an interior staircase to upper-floor rooms including an independent study and partitioned living spaces.

\paragraph{Laboratory.}
Base Agent produces a single-room lab centered on one island bench, a fume hood, and wall-side cabinets.
EnRealWorld constructs a partitioned multi-zone facility organized around a shared corridor that links distinct functional areas, including benchtop preparation bays, automated testing stations with rail-mounted manipulators, dedicated multi-tier wire storage racks, and corridor-connected zones navigated by mobile laboratory robots.

\subsection{Worlds from Paintings}
\label{sec:supp_f2}

\begin{figure*}[tp]
    \centering
    \includegraphics[width=\textwidth,height=.89\textheight,keepaspectratio]{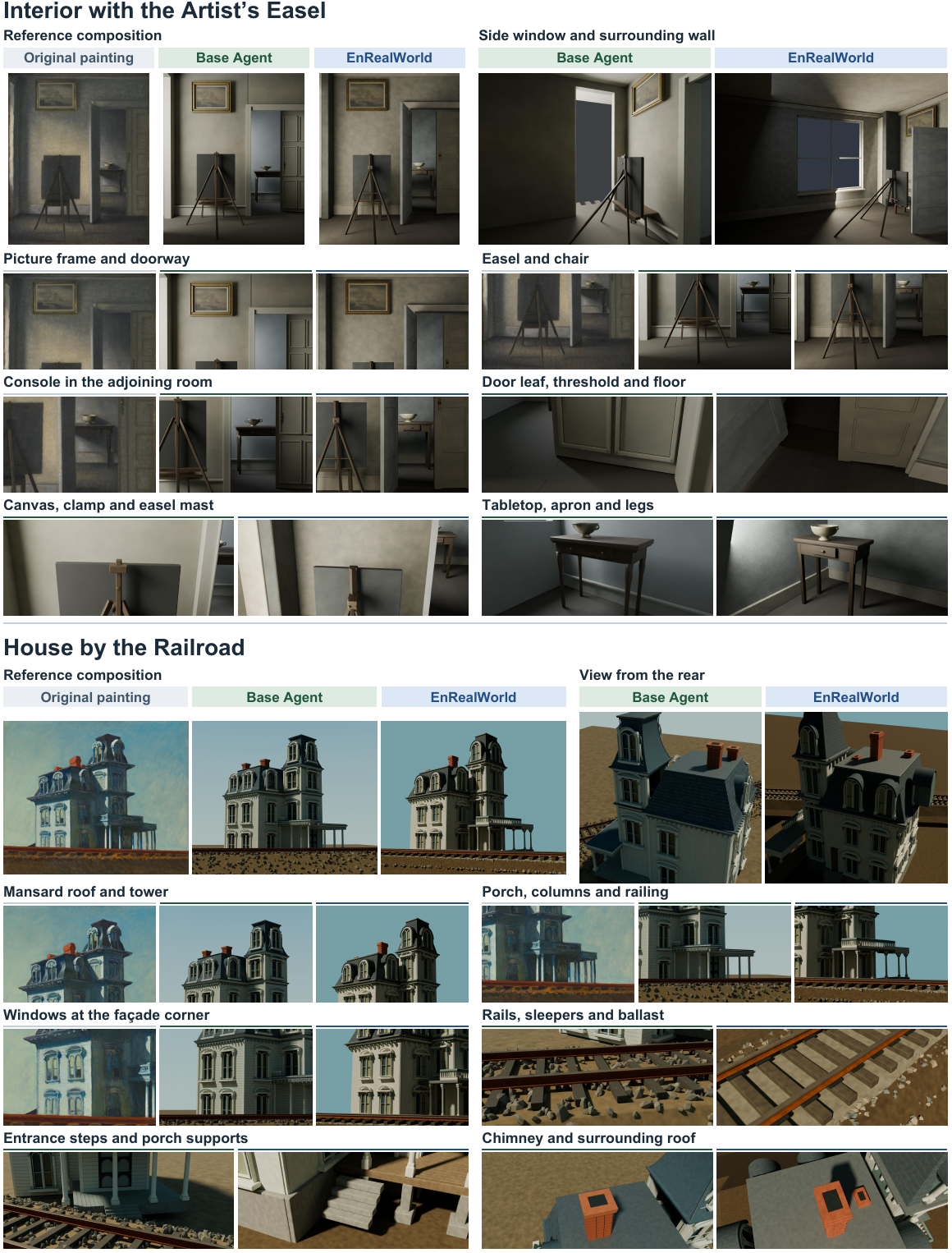}
    \caption{
    \textbf{Detailed comparisons for worlds reconstructed from paintings.} 
    For \emph{Interior with the Artist's Easel} (top) and \emph{House by the Railroad} (bottom), the reference-view panels compare the original painting with renders from Base Agent (green) and EnRealWorld (blue). 
    The novel views and structural close-ups compare the two methods directly.
    EnRealWorld improves both reference-view fidelity and off-camera structural completion.
    }
    \label{fig:supp_f2}
\end{figure*}

\cref{fig:supp_f2} presents multi-view and close-up comparisons for the two painting-based scenes, \emph{Interior with the Artist's Easel} and \emph{House by the Railroad}, pairing reference-view crops against the original paintings alongside unseen viewpoints and structural close-ups.
Across both scenes, EnRealWorld matches the reference composition and material tone more faithfully while completing off-camera architecture and resolving fine contact and joinery details.

\paragraph{Interior with the Artist's Easel.}
In the reference composition, both methods recover the framed painting, foreground easel, and adjoining room, while EnRealWorld better preserves the muted plaster patina, doorway proportions, and background furniture, including the chair beside the console table visible through the doorway.
Outside the reference view, Base Agent leaves the side wall with an unframed full-height cutout opening into empty space, whereas EnRealWorld completes a recessed multi-pane side window with casing, sill, and raking daylight across the floor.
Close-ups show cleaner structural and contact details: a continuous floor transition and a properly proportioned open paneled door leaf at the threshold, a seated rear mast and upper clamp gripping the canvas edge, and a console table with an explicit drawer front, knob, clean leg-to-apron joinery, and a bowl resting squarely on the tabletop.

\paragraph{House by the Railroad.}
In the reference view, EnRealWorld aligns more closely with the painting's tonal contrast, mansard roof silhouette, corner window lintels, and two-level porch proportions.
From the unseen rear viewpoint, Base Agent leaves the back mansard roof plain, while EnRealWorld extends dormers, tall windows, and consistent cornice trim across the rear facade.
In local close-ups, EnRealWorld models recessed multi-pane corner windows with arched hood moldings, supports the porch columns on entrance steps and masonry plinths, integrates stepped brick chimney caps into the dormer roofs, and replaces Base Agent's scattered coarse rocks with evenly spaced sleepers, continuous rusted rails, and a graded ballast bed.

\subsection{Worlds from Architectural Drawings}
\label{sec:supp_f3}

\begin{figure*}[tp]
    \centering
    \includegraphics[width=\textwidth,height=.89\textheight,keepaspectratio]{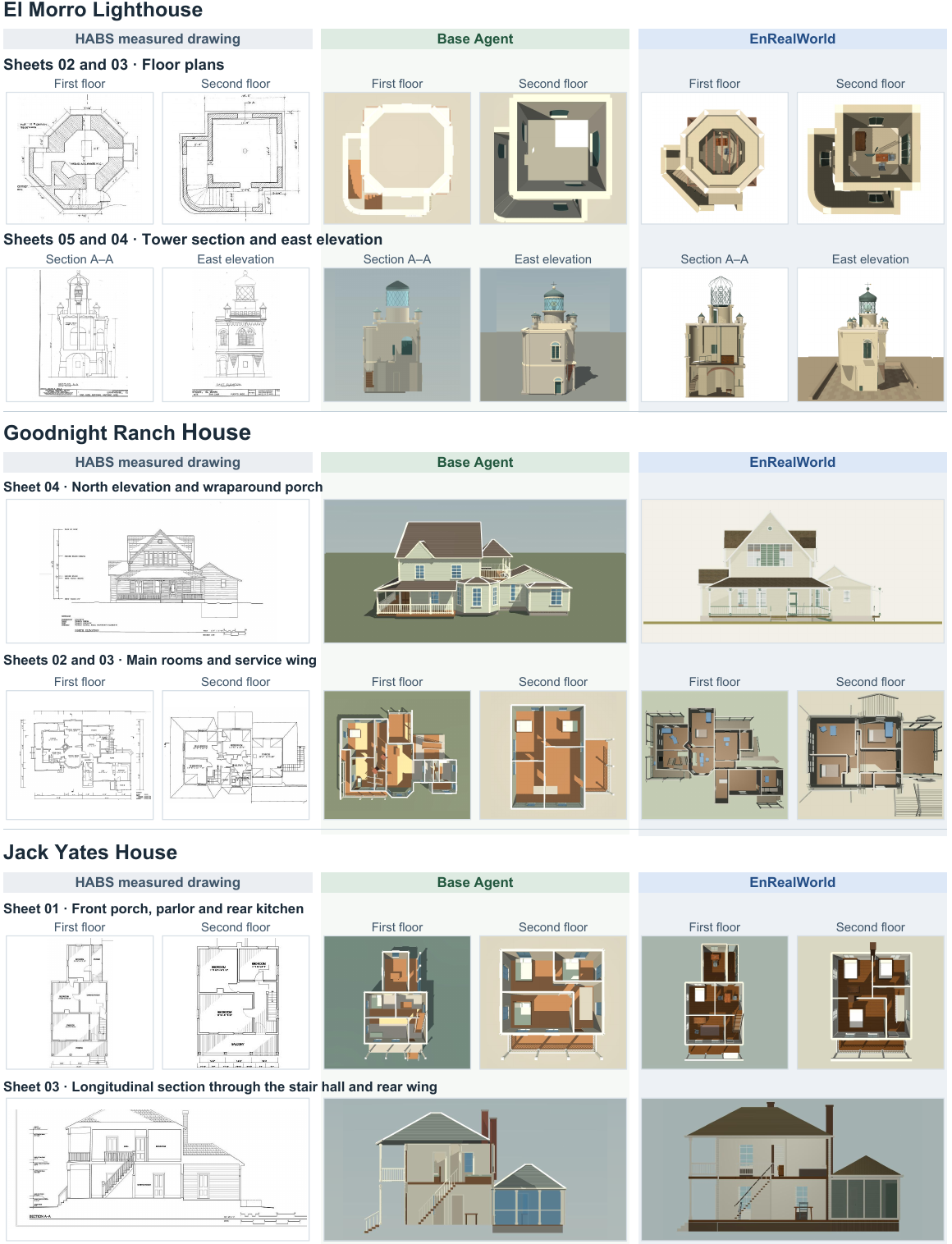}
    \caption{
    \textbf{Detailed comparisons for worlds reconstructed from HABS architectural drawings.} 
    For \emph{El Morro Lighthouse} (top), \emph{Goodnight Ranch House} (middle), and \emph{Jack Yates House} (bottom), each row aligns the original HABS measured drawing (left) with orthographic renders from Base Agent (center, green) and EnRealWorld (right, blue) across floor plans, elevations, and vertical sections. 
    EnRealWorld achieves higher fidelity in both multi-story interior layouts and cross-sectional structural consistency.
    }
    \label{fig:supp_f3}
\end{figure*}

\cref{fig:supp_f3} compares reconstructions from Historic American Buildings Survey (HABS) measured drawings across \emph{El Morro Lighthouse}, \emph{Goodnight Ranch House}, and \emph{Jack Yates House}, aligning orthographic floor plans, exterior elevations, and vertical sections directly against the original HABS sheets.
Across all three buildings, EnRealWorld maintains stronger cross-sheet consistency between exterior massing, multi-story room layouts, and interior vertical sections than Base Agent.

\begin{figure*}[tp]
    \centering
    \includegraphics[width=\textwidth,height=.88\textheight,keepaspectratio]{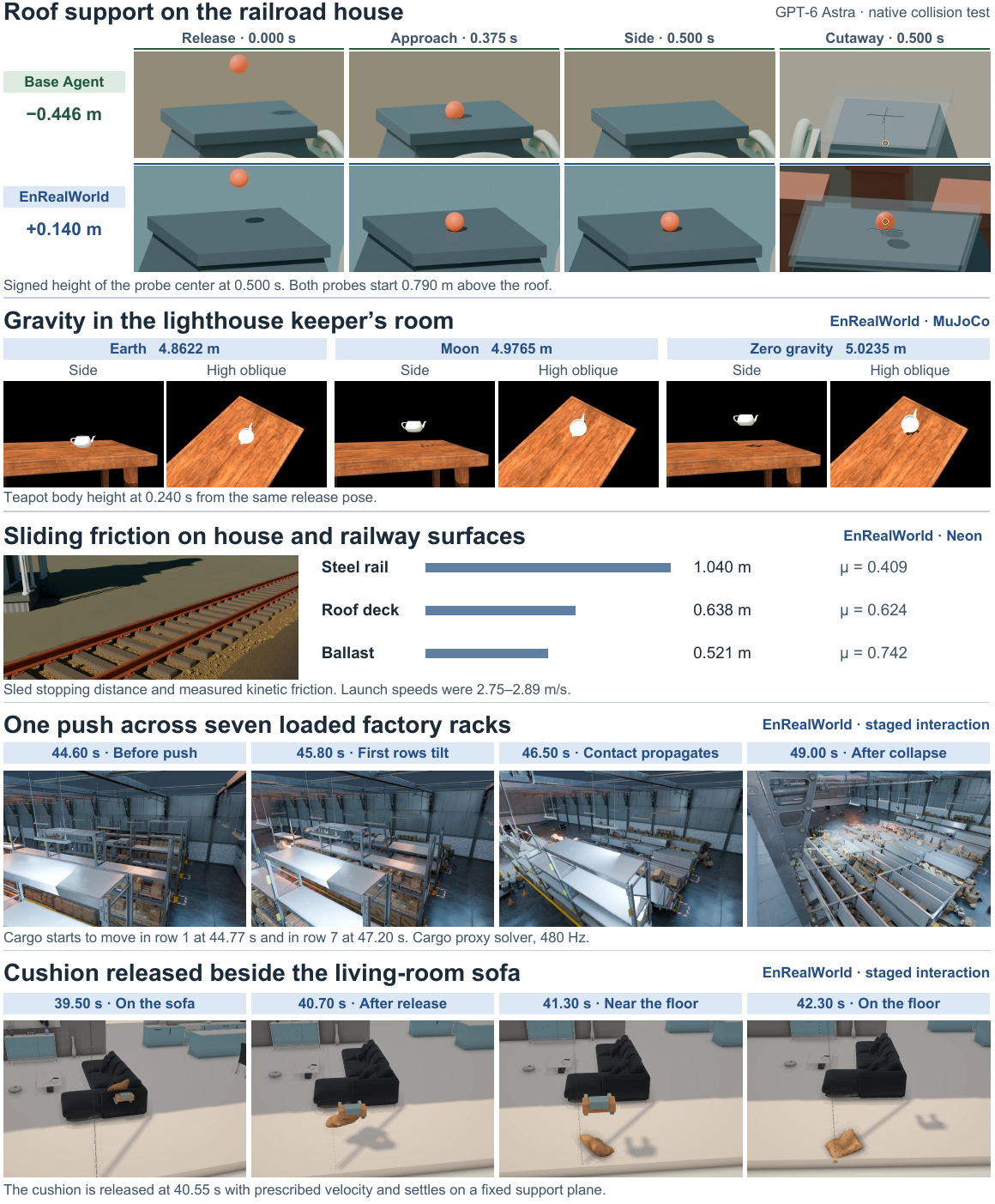}
    \caption{
    \textbf{Controlled physical interventions across reconstructed worlds.}
    From top to bottom: 
    (1)~native roof-support collision test on \emph{House by the Railroad}, where a probe passes through Base Agent's roof but rests on EnRealWorld's roof,
    (2)~teapot drop under Earth, Moon, and zero gravity in the lighthouse keeper's room, (3)~sled stopping distances and measured kinetic friction across steel rail, roof deck, and ballast surfaces,
    (4)~chain collapse across seven loaded factory storage racks triggered by a single push,
    and (5)~release, descent, and floor settling of a deformable living-room cushion.
    }
    \label{fig:supp_g1}
\end{figure*}

\paragraph{El Morro Lighthouse.}
On the first-floor plan, Base Agent leaves the octagonal interior largely hollow, omitting the concentric inner masonry core and stepped stairwell partition specified in the HABS plan, whereas EnRealWorld accurately recovers the thick octagonal masonry walls, entry stair, and inner chamber geometry.
On the second floor, EnRealWorld aligns the perimeter stairwell wrap, wall openings, and furnished keeper's room with the measured plan.
The vertical section (\emph{Section A--A}) makes the structural gap especially clear.
Base Agent's section remains largely solid and fails to expose the multi-level interior chambers, whereas EnRealWorld reveals the vaulted ground chamber, ascending staircase, second-floor keeper's room, and glazed lantern dome consistently with the HABS section and exterior elevation.

\paragraph{Goodnight Ranch House.}
On the north elevation, EnRealWorld matches the centered front gable, grouped second-story windows, full-width front porch with railings, and lower right-hand service wing shown in the HABS sheet, whereas Base Agent produces an inconsistent roof orientation and massing with an off-center dormer and protruding bay window.
Across both floor plans, Base Agent compresses the footprint and reduces the second floor to a simplified rectangular layout, while EnRealWorld preserves the irregular main-house and service-wing geometry, wraparound porches, hallway and stair circulation, and room partitions across both stories.

\paragraph{Jack Yates House.}
On the floor plans, EnRealWorld adheres more closely to the HABS room divisions across the front porch, parlor, dining room, right-side stair hall, rear kitchen wing, and three second-floor bedrooms.
In the longitudinal section through the stair hall and rear wing, Base Agent introduces duplicated rear chimneys and an irregular upper-hall partition with misaligned stair geometry, whereas EnRealWorld reproduces the single rear chimney, hipped roof proportions, continuous staircase to the upper landing, aligned windows, and lower rear kitchen wing in close agreement with the measured section.

\section{Controlled Physical Interventions}
\label{sec:supp_g1}

Complementing the studio drop and door-collision tests (\cref{fig:physical_responses}), \cref{fig:supp_g1} evaluates physical executability under five additional controlled interventions.

\paragraph{Roof support collision.}
In the GPT-6 Astra reconstruction of \emph{House by the Railroad}, a spherical probe is released $0.790\,\text{m}$ above the visible roof surface.
By $0.500\,\text{s}$, the probe passes directly through Base Agent's roof to a signed center height of $-0.446\,\text{m}$, as confirmed by the cutaway view, whereas EnRealWorld's roof catches and supports the probe cleanly at $+0.140\,\text{m}$.

\paragraph{Gravity response.}
In EnRealWorld's lighthouse keeper's room executed in MuJoCo, releasing a teapot from the same initial pose produces distinct heights at $0.240\,\text{s}$ across gravity regimes.
Under Earth gravity the teapot rests on the tabletop at $4.8622\,\text{m}$, under lunar gravity it is mid-descent at $4.9765\,\text{m}$, and under zero gravity it remains suspended at $5.0235\,\text{m}$.

\paragraph{Surface-dependent sliding friction.}
Launching a sled at $2.75\text{--}2.89\,\text{m/s}$ across three surfaces in EnRealWorld's \emph{House by the Railroad} scene with Gemini 4 Neon yields physically ordered stopping distances and kinetic friction coefficients: $1.040\,\text{m}$ ($\mu = 0.409$) on the steel rail, $0.638\,\text{m}$ ($\mu = 0.624$) on the roof deck, and $0.521\,\text{m}$ ($\mu = 0.742$) on the gravel ballast.

\paragraph{Staged multi-body and deformable interactions.}
We further verify long-horizon contact propagation and deformable landing in the factory and home scenes.
In the factory warehouse simulated at $480\,\text{Hz}$ with a cargo proxy solver, a single lateral push on the first of seven loaded storage racks initiates a sequential collapse across all rows, with cargo motion starting in row~1 at $44.77\,\text{s}$ and reaching row~7 by $47.20\,\text{s}$.
In the living-room scene, a soft cushion lifted from the sofa by an actuator and released at $40.55\,\text{s}$ falls freely, deforms on impact, and settles stably on the floor support plane by $42.30\,\text{s}$.


\end{document}